\documentclass[lettersize,journal]{IEEEtran}
\pdfoutput=1
\usepackage{amsmath,amsfonts}
\usepackage{array}
\usepackage[caption=false,font=normalsize,labelfont=sf,textfont=sf]{subfig}
\usepackage{textcomp}
\usepackage{stfloats}
\usepackage{url}
\usepackage{verbatim}
\usepackage{graphicx}
\usepackage{cite}
\usepackage{relsize}
\usepackage{mathtools}
\usepackage{array,tabularx}
\usepackage{xcolor}
\usepackage{booktabs,arydshln}
\usepackage{multirow}
\usepackage{enumitem}
\usepackage{tikz}
\usepackage{threeparttable}
\usepackage{url}
\usepackage[ruled,vlined,linesnumbered]{algorithm2e}
\usepackage[noend]{algpseudocode}
\usepackage{amssymb}
\usepackage{cleveref}

\begin{document}

\title{Parse Trees Guided LLM Prompt Compression}

\author{Wenhao Mao, Chengbin Hou*, Tianyu Zhang, Xinyu Lin, Ke Tang, \IEEEmembership{Fellow~IEEE}, Hairong Lv*
\thanks{Wenhao Mao, Tianyu Zhang, and Hairong Lv are with Ministry of Education Key Laboratory of Bioinformatics, BNRIST Bioinformatics Division, Department of Automation, Tsinghua University, Beijing, China.}
\thanks{Chengbin Hou and Xinyu Lin are with School of Computing and Artificial Intelligence, Fuyao University of Science and Technology, Fuzhou, China.}
\thanks{Ke Tang is with Department of Computer Science and Engineering, Southern University of Science and Technology, Shenzhen, China.}
\thanks{* Corresponding authors: Chengbin Hou; Hailong Lv}
\thanks{E-mail: houcb@fyust.edu.cn; lvhairong@tsinghua.edu.cn}
\thanks{Note: This is the author's version of the work accepted for publication in IEEE Transactions on Pattern Analysis and Machine Intelligence. The definitive version will be published by IEEE.}
% \thanks{Submitted to IEEE TPAMI}
% \thanks{Manuscript received xxxx xx, 2024; revised xxxx xx, 2024}
}

% \markboth{Journal of \LaTeX\ Class Files,~Vol.~14, No.~8, August~2021}

\maketitle

\begin{abstract}
Offering rich contexts to Large Language Models (LLMs) has shown to boost the performance in various tasks, but the resulting longer prompt would increase the computational cost and might exceed the input limit of LLMs. 
Recently, some prompt compression methods have been suggested to shorten the length of prompts by using language models to generate shorter prompts or by developing computational models to select important parts of original prompt.
The generative compression methods would suffer from issues like hallucination, while the selective compression methods have not involved linguistic rules and overlook the global structure of prompt. 
To this end, we propose a novel selective compression method called PartPrompt. It first obtains a parse tree for each sentence based on linguistic rules, and calculates local information entropy for each node in a parse tree. These local parse trees are then organized into a global tree according to the hierarchical structure such as the dependency of sentences, paragraphs, and sections. After that, the root-ward propagation and leaf-ward propagation are proposed to adjust node values over the global tree. Finally, a recursive algorithm is developed to prune the global tree based on the adjusted node values.
The experiments show that PartPrompt receives the state-of-the-art performance across various datasets, metrics, compression ratios, and target LLMs for inference. The in-depth ablation studies confirm the effectiveness of designs in PartPrompt, and other additional experiments also demonstrate its superiority in terms of the coherence of compressed prompts and in the extreme long prompt scenario.

\end{abstract}

\begin{IEEEkeywords}
Large Language Models, Prompt Compression, Parse Trees, Prompt Structure, Text Pattern Analysis.
\end{IEEEkeywords}

\section{Introduction}
\IEEEPARstart{L}{arge} Language Models (LLMs) have achieved remarkable performance on various tasks such as question answering, summarization, multimodal generation, and information extraction \cite{achiam2023gpt,meta2024llama3}. Prompting LLMs with adequate task-related contexts can enhance their performance. The prompting techniques, like in-context learning \cite{dong2022survey}, chain-of-thought \cite{wei2022chain}, and retrieval augmented generation \cite{lewis2020retrieval}, have shown noticeable performance gains for answering questions that require long-tail knowledge \cite{li2024role} and reasoning ability \cite{fu2023chainofthoughthubcontinuouseffort}.

Most LLMs adopt the transformer architecture, which results in the computational cost of LLMs being proportional to the square of input context length \cite{vaswani2017attention}. In addition, there is an upper limit to the number of tokens that can be processed by LLMs \cite{vaswani2017attention}. The use of long prompt for providing rich contexts to LLMs would also significantly increase the computational cost and may exceed the input token limit. To this end, prompt compression, by shortening the length of prompt, is suggested to reduce the computational cost during LLM inference and make LLMs possible for handling prompts beyond the input token limit (or indirectly increasing the limit).

Recently, some prompt compression methods \cite{chuang2024learning,gilbert2023semantic} feed the original prompt into a language model for generating the compressed prompt. These generative compression methods exploit the language understanding and generation ability of language models, but would encounter issues like hallucination \cite{ji2023survey} due to the limitations of generative language models. Another typical type of prompt compression methods \cite{li2023compressingcontextenhanceinference,jiang2023llmlingua,pan2024llmlingua,jung2024discrete,huang2023boosting} obtain the compressed prompt by selecting a portion of the original prompt without generating new contents, and thus alleviate the hallucination issue. These selective compression methods typically adopt a measure to evaluate the importance of tokens in the original prompt, and preserve the important parts while outputting the compressed prompt.

Regarding existing selective prompt compression methods, the parts of the original prompt to remain or remove are determined mainly by a computational model, e.g., using a small language model to calculate information entropy as the measure. The computational models have not yet involved the linguistic rules, which have been previously shown effective in various learning tasks 
\cite{zhang2020sg,hong2019learning,cao2019interpretable,unno2006trimming}. 
Besides, the computational cost of prompt compression methods is also deserved to consider, since one fundamental motivation of prompt compression is to reduce the computational cost for LLM inference \cite{li2023compressingcontextenhanceinference,jiang2023llmlingua}. 
Moreover, the streaming processing of original prompt by compression methods may overlook the connecting patterns among sentences and the hierarchical structure of the whole prompt, which we refer as the human writing logic especially while writing long prompts. 

To address these challenges, this work proposes to leverage \underline{Par}se \underline{t}rees to guide the \underline{Prompt} compressing process (namely PartPrompt) incorporating with the local information entropy and the global prompt patterns. Specifically, PartPrompt first analyzes a parse tree for each sentence and obtains the local information entropy of tokens within each sentence. Second, a global tree is constructed to reflect connecting patterns among sentences and hierarchical structure of the whole prompt. Third, the node value based on the entropy is adjusted over the global tree by the newly proposed root-ward propagation and leaf-ward propagation. And finally, a recursive algorithm is developed to prune the global tree based on the adjusted node value. It is noted that, during prompt compression, the linguistic rules are introduced by the parse trees of sentences; the computational cost is reduced by the local approximated entropy; the human writing logic is considered in the construction of global tree and the adjustment of node value.

Extensive experiments are conducted on eight datasets to present the performance across various prompt task types (including understanding, summarization, in-context learning, and math question answering) and prompt text types (including English, non-English, non-natural language, short, and long prompts).
The results demonstrate that PartPrompt considerably outperforms the state-of-the-art prompt compression methods over several compression ratios and metrics. And the superior performance of PartPrompt can be also observed when feeding the compressed prompt to different LLMs as the target model for inference. Besides, in-depth ablation experiments confirm the effectiveness of the key components in PartPrompt such as introducing parse trees, constructing the global tree, and adjusting the node value. Moreover, we further explore the performance of PartPrompt under the extreme long prompt case, and investigate the coherence of the compressed prompt via quantitative metrics and intuitive examples. 

Apart from extensive experiments and in-depth analysis, the main technical contributions are as follows:
\begin{itemize}[leftmargin=*]
    \item This work introduces parse trees, a kind of linguistic rules and text patterns, to guide the prompt compression process for the first time in literature. For this purpose, we transform the prompt compression problem of selecting tokens into the tree constructing and pruning problem.
    \item To capture the global structure, we propose to construct a global tree for the original prompt based on the dependency of sentences, paragraphs, and sections. And we further propose the novel root-ward propagation and leaf-ward propagation to adjust the value of nodes in the tree. Both techniques are motivated by the human writing logic.
    \item Unlike sequentially evaluating and removing unimportant parts as previous methods done, a new recursive algorithm is developed to remove the unimportant parts of original prompt by pruning over the global tree. And with the help of the global tree, it becomes reasonable to adopt local approximated information entropy, thereby saving computational cost compared to using global information entropy.
    \item To conduct comprehensive experiments, we newly create one, re-crawl two, and re-clean one prompt compression benchmark datasets. To benefit future research, the source code along with all datasets are freely available at \url{https://github.com/LengendaryHippopotamus/PartPrompt}.
\end{itemize}

\section{Related Work}
\subsection{Large Language Models and Prompt Engineering}

Early language models were applied for tasks such as text classification and machine translation \cite{khurana2023natural,chowdhary2020natural}. Subsequently, some works \cite{vaswani2017attention,radford2018improving,kaplan2020scaling} discovered that the performance of language models is highly correlated to their parameter scales, leading to development of LLMs in pursuit of more powerful capabilities \cite{radford2019language,brown2020language}. After that, many LLMs with billions of parameters \cite{achiam2023gpt,meta2024llama3} were developed and have achieved remarkable success in a wide variety of tasks.

The recent LLMs have also emerged many new abilities with the help of prompts \cite{wei2022emergent} such as in-context learning \cite{dong2022survey}, chain-of-thought \cite{wei2022chain}, and retrieve augmented generative \cite{lewis2020retrieval}. A multitude of works then attempt to enhance the performance of LLMs 
via prompt engineering. Some of them manually design prompts \cite{wei2022chain}, whereas other works focus on designing prompts in a more automatic way \cite{zhang2022automatic}.

Prompting LLMs with adequate task-related contexts can enhance the ability of LLMs. However, the prompt techniques such as \cite{dong2022survey,wei2022chain,lewis2020retrieval} often require a relatively longer prompt, which would considerably increase the computational cost during inference due to the transformer architecture. To tackle this issue, recent works have begun to explore prompt compression, i.e., using the compressed prompt to replace the given prompt while preserving performance. And this work also intends to investigate the prompt compression problem.

\subsection{Prompt Compression} 

The prompt compression methods can be divided into two categories: generative compression and selective compression. Generative compression utilizes a language model to take the original given prompt as the input, which is then asked to generate the compressed prompt typically from the same language model.

Specifically, \cite{gilbert2023semantic} directly employs LLMs to compress the given prompt, and this work then analyzes the ability of LLMs in retaining semantics and understanding contents after compression; \cite{chuang2024learning} trains an encapsulation model with a semantic loss and a reward function, and the trained model is then adopted to generate the compressed prompt.

Selective compression, on the other hand, selects a portion of the original prompt as the compressed prompt, which can better preserve the original contents and avoid hallucinations.

Some selective compression methods directly employ an existing pretrained language model to evaluate the importance of tokens. Selective-Context \cite{li2023compressingcontextenhanceinference} obtains the compressed prompt by retaining tokens with higher information entropy, which is computed by a pretrained language model. Then, LLMLingua \cite{jiang2023llmlingua} further introduces the budget controller, iterative token-level compression, and distribution alignment for a better information entropy estimation and a higher compression rate. Later, LongLLMLingua \cite{jiang2023longllmlingua} additionally employs a question-aware approach. Apart from them, other selective compression methods try to train a new model to evaluate the importance of tokens. LLMLingua2\cite{pan2024llmlingua} trains a model to learn the token compression given by the GPT4. \cite{jung2024discrete} exploits reinforcement learning to train a model for deciding which tokens to remove, and \cite{huang2023boosting} also uses reinforcement learning to train a model for pruning the prompt with the chain-of-thought style.

Distinguished from previous selective compression works, this work tries to transform the prompt compression problem of selecting tokens as the tree pruning problem. During such transformation, we consider linguistic rules, hierarchical structure of prompts, and human writing logic, while building the tree and updating the values of nodes for tree pruning.

\subsection{Text Pattern Analysis and Compression}
The textual data contains latent patterns that can be adopted to assist text or language analysis and processing. For example, \cite{li2021text} employs backbone information to help the transformer model to learn the text encoding and representation.
Some explicit text patterns, e.g., the patterns given by linguistic parse trees, can be easily understood by humans and have been applied to many machine learning tasks. For instances, \cite{zhang2020sg} employs the parse tree to aid natural language processing tasks like machine translation. 
Both \cite{hong2019learning} and \cite{cao2019interpretable} employ parse trees to enhance the reasoning ability of models.
Note that, the most related work to our work, \cite{unno2006trimming} also utilizes the explicit text patterns, i.e, parse trees, to compress sentences.

Unlike the sentence compression, the prompt compression for LLMs has its own challenges. First, the prompt often consists of multiple sentences or even multiple paragraphs and sections, which involves much more abundant patterns. Second, the compressed output is not only aiming for human understanding but also for assisting LLMs in various tasks. Also note that, the patterns in LLM prompt, such as linguistic parse trees and the hierarchical structure of sentences,  have not yet been explored in the prompt compression problem.

\begin{figure*}[hbp]
    \centering
    \includegraphics[width=0.95\textwidth]{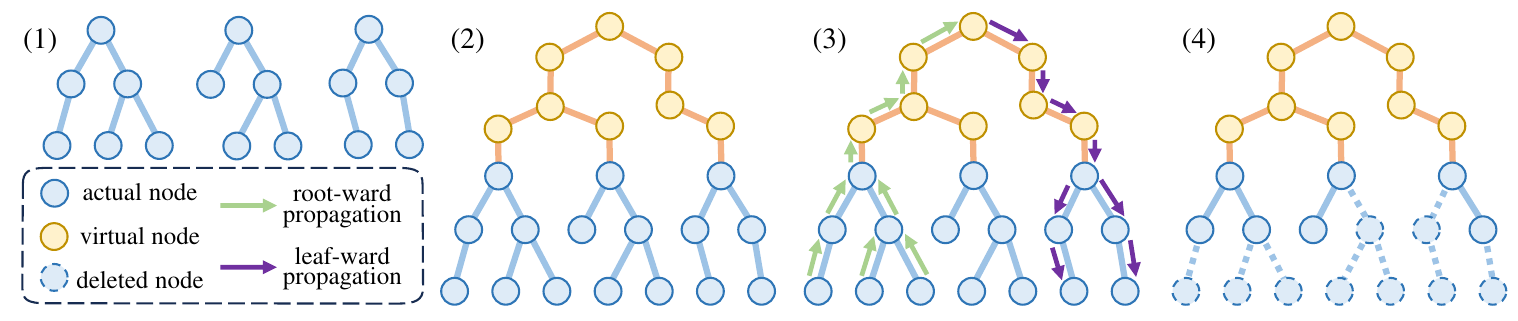}
    \caption{Overall framework of PartPrompt: (1) local parse trees for sentences, (2) global parse tree construction, (3) node value adjustment, (4) tree compression.}
    \label{fig1}
\end{figure*}

\section{Problem Formulation}
\noindent\textbf{Definition 1. Prompt} $R \triangleq r_1r_2\cdots r_n$ is a sequence of tokens, which is the natural language input to LLMs for performing a specific task. For each token $r_i$ in $R$, function $C(\cdot)$ calculates the length of a token, 
and function $E(\cdot)$ calculates the importance or value of a token.
Given a prompt $R$ containing $n$ tokens, the length of the prompt can be obtained by $C(R) \triangleq \sum_{i=1}^n C(r_i)$, and the value of the prompt can be obtained by $E(R) \triangleq \sum_{i=1}^n E(r_i)$.

\noindent\textbf{Definition 2. Selective Compression} refers to the compressed prompt $R_{\text{cp}} \triangleq r_{w_1}r_{w_2}\cdots r_{w_l}$ being selected from a part of the original prompt $R=r_1r_2\cdots r_n$, where $\{w_1,w_2,\cdots, w_l\}$ is a subset of $\{1,2,\cdots,n\}$ satisfying $w_1<w_2<\cdots <w_l$, which indicates that the selective compression does not generate new tokens, and preserves the original order. Accordingly, the compression ratio between the compressed prompt and original prompt can be defined by
\begin{equation}
    \tau \triangleq \frac{C(R_{\text{cp}})}{C(R)}.
\end{equation}

\noindent\textbf{Definition 3. Selective Prompt Compression Problem} can be formulated as: for a given prompt $R$ and compression ratio $\tau$ where $0<\tau<1$, find the compressed prompt $R_{\text{cp}}$, selecting from the original $R$, such that the overall important score or value of $R_{\text{cp}}$ is maximized without exceeding the upper token limit set by $\tau$ of $R$ during the compression process, i.e., 
\begin{equation}
    \max E(R_{\text{cp}})~~\text{s.t.}~C(R_{\text{cp}})\leq \tau C(R).
\end{equation}

\noindent\textbf{Definition 4. Parse Tree} $T \triangleq (R,V,v_\text{root},f)$ consists of four terms, where $R$ is the given prompt; $V=\{v_1,v_2,\cdots,v_n\}$ is a node set for the parse tree wherein $v_\text{root}$ is the root node without parent node; for each node $v_i\in V$, function $f:V\setminus \{v_\text{root}\} \mapsto V$ maps each node to its parent node except for the root node. 
Note that, each node $v_i$ corresponds one-to-one to a token $r_i$, and hence to obey the order of tokens in the prompt, the nodes in a parse tree also retains the same order. We define $v_{w_1}<v_{w_2}$ (here $<$ indicates that it is ordered earlier in the sequence among sibling nodes)
for subscript $w_1<w_2$. The node with the smallest index among child nodes is called the first child node. The length and value of a node can be defined by the corresponding token, i.e., $C(v_i) \triangleq C(r_i)$ and $E(v_i) \triangleq E(r_i)$ respectively.

Remarks: We expand the definition of the parse tree based on the foundation of traditional linguistics. If the given prompt contains only one sentence, the parse tree is exactly the traditional linguistic parse tree of the sentence. If the given prompt includes multiple sentences, the parse tree then becomes the global parse tree, which integrates multiple linguistic (also referred as local) parse trees following some logic.

\noindent\textbf{Definition 5. Subtree} $T_j=(R_j,V_j,v_{\text{root},j},f_j)$ is a portion of a given parse tree $T=(R,V,v_\text{root},f)$. The node $v_{\text{root},j}$, a particular node in the node set $V$ of given parse tree $T$, is the root node of the subtree. And it, along with all its child nodes, constitutes the set $V_j$.

\noindent\textbf{Definition 6. Compressed Tree} $T_{\text{cp}} \triangleq (R_{\text{cp}},V_{\text{cp}},v_{\text{root,cp}},f_{\text{cp}})$ is the remaining part of the given parse tree $T=(R,V,v_\text{root},f)$ after pruning some subtrees $T_1, T_2,\cdots T_l$.
Let $k$-th pruned subtree be $T_k=(R_k,V_k,v_{\text{root},k},f_k)$ for $k\in \{1,2,\cdots,l\}$, then the deleted node set is $\bigcup_{k=1}^lV_k$, the retained node set is $V_{\text{cp}}=V\setminus (\bigcup_{k=1}^lV_k)$. 
Since all nodes in the retained node set are also the nodes in the given tree $T$, the compressed prompt $R_{\text{cp}}$ is a selective compression of $R$.

\section{The Proposed Method}
This section elaborates the proposed method in detail, and the overall framework of proposed PartPrompt is illustrated in Figure \ref{fig1}. To be more specific, for a given prompt $R$, we first slice it into sentences $[R_1,R_2,\cdots,R_m]$ and calculate the information entropy of each token within each sentence. A parse tree is then built based on linguistic rules for each sentence so that we obtain corresponding $m$ local parse trees $[T_1,T_2,\cdots,T_m]$. 
Second, a global parse tree $T$ is constructed by using virtual nodes to merge these local parse trees based on the structure of prompt. 
Third, a node value adjustment module, including root-ward propagation and leaf-ward propagation, is proposed to adjust the original value of each node based on the human writing logic.
Finally, a recursive algorithm is developed to prune the global tree, which yields the compressed tree $T_{cp}$ and accordingly the compressed prompt $R_{cp}$ for a given compress ratio $\tau$.

\subsection{Information Entropy Approximation}
\label{InformationEntropyApproximation}
Given a prompt $R$ that contains multiple sentences, we slice it into a list of sentences $[R_1,R_2,\cdots R_j,\cdots R_m]$. For each sentence $R_j$, it can further be divided into a sequence of tokens 
$r_{j,1}r_{j,2}\cdots r_{j,i}\cdots r_{j,n_j}$. The conditional probability of a token $r_{j,i}$ given its preceding token sequence can be computed by
\begin{equation}
    p(~r_{j,i}~|~r_{j,<i},r_{<j}),
\end{equation}
where the subscript $j$ denotes sentence index and $<j$ indicates all the tokens prior to sentence $j$, while the subscript $i$ denotes token index and $<i$ indicates all the tokens prior to the token $i$ within the sentence $j$. The corresponding information entropy \cite{shannon1948mathematical} of a token $r_{j,i}$ then becomes
\begin{equation}
    E(r_{j,i})=-\log ~p(~r_{j,i}~|~r_{j,<i},r_{<j}).
    \label{eq4}
\end{equation}

As shown in Equation (\ref{eq4}), to obtain the information entropy of a token, we can compute its conditional probability, which is exactly what the language model does. Accordingly, the information entropy of a token $r_{j,i}$ estimated by a language model can be calculated by
\begin{equation}
    -\log~p_{\text{LM}}(~r_{j,i}~|~r_{j,<i},r_{<j}).
\end{equation}

In order to reduce the computational cost of calculating information entropy, we make the following approximation
\begin{equation}
p(~r_{j,i}~|~r_{j,<i},r_{<j})\approx p(~r_{j,i}~|~r_{j,<i}),
\end{equation}
where the term $r_{<j}$ is removed, which means the information entropy of a token is only considered within its own corresponding sentence. Consequently, the information entropy calculated by this approximation finally becomes
\begin{equation}
    E_\text{LM}(r_{j,i})=-\log~p_{\text{LM}}(~r_{j,i}~|~r_{j,<i}).
    \label{eq8}
\end{equation}

Regarding the information entropy approximation in Equation (\ref{eq8}), it is worth noting that, the computational cost is saved especially for a long prompt containing multiple sentences. Besides, to tackle the potential performance drop caused by the information entropy approximation and make the approximation more promising, we suggest to consider the global structure of a given prompt using a well-organized tree and adjust the approximated information entropy based on the tree, which are presented in the following sections.

\subsection{Global Parse Tree Construction}\label{GlobalParseTree}

A local parse tree, based on linguistic rules, is a tree that analyzes word dependencies for a given sentence. Concretely, in a local parse tree $T_j$ corresponding to a sentence $R_j$, the word $r_i$, attached to the child node $v_i$, is considered dependent on the word of its parent node via mapping $f_j$ on $v_i$. The word of global verb in this sentence is located at the root node $v_{\text{root},j}$ without its parent dependency. We adopt Stanford NLP toolkit \cite{manning2014stanford} to build all local parse trees $[T_1,T_2,\cdots,T_m]$ for all sentence $[R_1,R_2,\cdots,R_m]$.

For the series of local parse trees $T_j=(R_j,V_j,v_{\text{root},j},f_j)$ for $j \in \{1,2\cdots m\}$, we introduce a new virtual node $\tilde{v}$ and make $v_{\text{root},1},v_{\text{root},2},\cdots, v_{\text{root},m}$ as its child nodes. In this way, these trees are aggregated into a single tree, which is referred as the global parse tree, denoted by $T$. For the virtual node $\tilde{v}$ introduced to aggregate trees, there is no actual token attached to the virtual node, and the initial value or information entropy is zero. In contrast, the nodes in local parse trees contain actual token(s) and are referred as actual nodes.

The aforementioned global tree connects all local parse trees into a global tree, but cannot reflect the hierarchical structure of the entire given prompt. Considering a common sentence-paragraph-section-document writing style, we suggest to carry out the aforementioned aggregation process in several steps. First, the virtual sentence nodes are created for each local parse tree corresponding to each sentence. Second, the virtual paragraph nodes are created to aggregate their subtree(s) belonging to the same paragraph. Third, the virtual section nodes are created to aggregate their subtree(s) belonging to the same section. Fourth, a virtual document node is created to aggregate all subtree(s) for the entire prompt.

The global parse tree $T$ has dependencies between child nodes and parent nodes: actual nodes (from local parse trees) depend on the corresponding virtual sentence nodes; virtual sentence nodes depend on virtual paragraph nodes; virtual paragraph nodes depend on virtual section nodes; virtual section nodes depend on the virtual document node (i.e., the root node of global tree $T$) for the entire given prompt. This hierarchical dependency relationship clearly reflects the hierarchical structure of the entire prompt. 

\subsection{Token Alignment}
\label{TokenAlignment}
For the actual nodes in the local parse trees analyzed by linguistic rules, there is a token attached to each actual node. However, the token in an actual node might not be one-to-one correspondence to the token analyzed by an LLM tokenizer, which is utilized in obtaining the value of an actual node to reflect its importance by a small well-trained LLM. Consequently, a token alignment module is needed to compute the value and length of the token in each actual node. 

The aim of token alignment is to ensure the smallest token retaining sufficient semantic integrity. To this end, we choose the tokenizer of establishing local parse trees (denoted as parse tree tokenizer) as the base, and utilize a small well-trained LLM with LLM tokenizer to compute the information entropy of the token(s) attached to actual nodes in local parse trees. The alignment between these two tokenizers is achieved through a rule-based matching algorithm, yielding the aligned information entropy or value $E_{\text{Aligned}}(v_i)$ and its length $C(v_i)$ for each node $v_i$ in local parse trees.

\begin{figure}[htbp]
    \centering
    \includegraphics[width=0.485\textwidth]{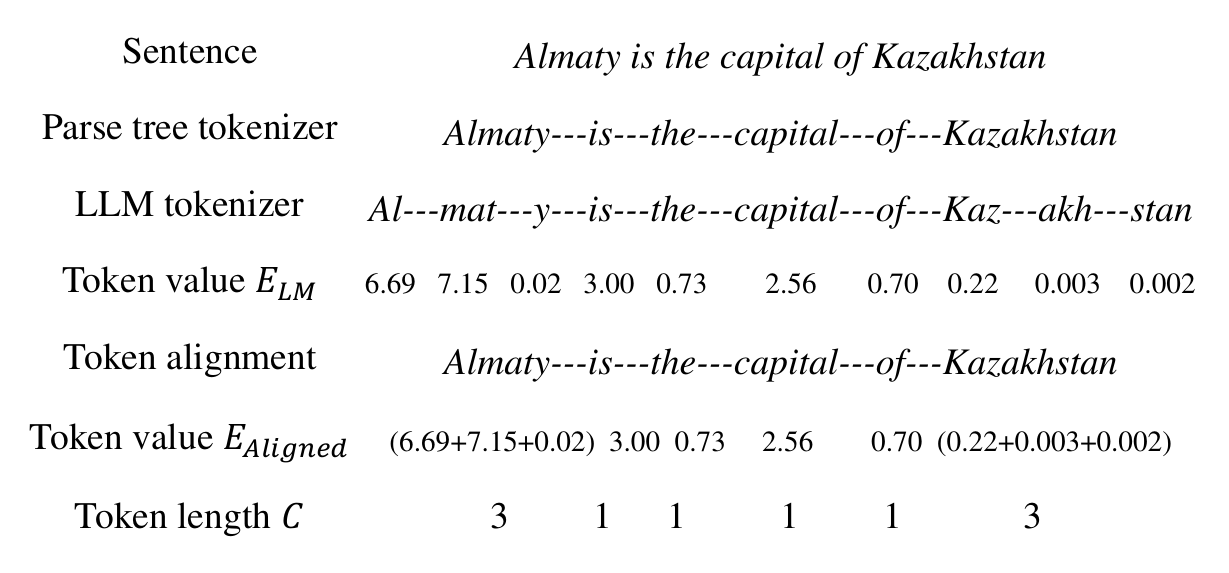}
    \caption{An example of token alignment.}
    \label{TokenAlignExample}
\end{figure}
Figure \ref{TokenAlignExample} illustrates a toy example. The token \textit{Almaty} is recognized as a single unit by the parse tree tokenizer, whereas the LLM tokenizer divides it into three distinct tokens: \textit{Al}, \textit{mat}, and \textit{y}. The information entropy $E_{\text{LM}}$ computed by LLM for the three tokens are $6.69$, $7.15$, and $0.02$. To maintain the semantic integrity, we adhere to make the parse tree tokenizer as the base, treating \textit{Almaty} as a unified token. Accordingly, the aligned information entropy $E_{\text{aligned}}$ of this unified token is the sum of the information entropy of the three tokens by LLM tokenizer, i.e., $(6.69 + 7.15 + 0.02)$. And the length of this unified token $C(\textit{Almaty})$ attached to the node is 3.

\subsection{Node Value Adjustment}
\label{InformationEntropyAdjustment}

Section \ref{GlobalParseTree} have introduced a novel global tree composing of actual nodes and virtual nodes. The hierarchical structure of virtual nodes simulates the typical writing style of sentence-paragraph-section-document. The purpose of virtual nodes is to harmonize the compression requests for the segments at the global scale, where each segment refers to a subtree rooted by a virtual node. For this purpose, two pivotal criteria have to consider. First, each virtual node should reflect the value of its corresponding segment. Second, the compression request for each segment should be able to propagate to its corresponding actual nodes as well as virtual nodes.

Given the aim and two criteria, a node value adjustment algorithm is developed to adjust the original values attached to the nodes of the global parse tree. It consists of two components: root-ward propagation and leaf-ward propagation. The pseudocode is presented in Algorithm \ref{AdjustmentAlgorithm}, where Lines 1-11 corresponds to the root-ward propagation and Lines 12-20 corresponds to the leaf-ward propagation.

Root-ward propagation tries to update the value of each virtual node (Lines 5 and 7) such that the value of a virtual node reflects the value of its corresponding segment. Inspired from the forward and backward propagation in neural networks, we employ a momentum-based approach to aggregate node values from leaf nodes to the root node of a segment, thereby assigning the averaged value to the corresponding virtual node. Note that, the value of actual nodes are not updated and we append the aggregated value to a list or vector $\vec{M}$, whereas the value of virtual nodes are updated recursively and we also append the aggregated value to $\vec{M}$.

Leaf-ward propagation attempts to update the value of each actual node (Lines 17 and 18) based on its original value and the compression request. The compression request is propagated recursively from the root node to leaf nodes. For a virtual node, we employ a scalar $M$ (at very beginning $M=1$) to cache its value, and we adjust $M$ with a hyper-parameter $a_2$ if it is the first virtual node at a hierarchy of the global tree. For an actual node, its adjusted value is obtained by adding its original value (i.e., the aligned information entropy) and the cached $M$ with an experiential adjustment hyper-parameter $a_1$. Note that, $a_1$ ensures the adjusted value retaining a notable distinction. And $a_2$ targets at emphasizing the value of the first part at each hierarchy of the global tree, such as the first sentence in a paragraph and the first paragraph in a section.

\begin{algorithm}
\caption{Node Value Adjustment}
\label{AdjustmentAlgorithm}
\KwIn{Global Parse Tree $T=(R,V,v_\text{root},f)$, Aligned Information Entropy $E_{\text{Aligned}}$, Hyper-parameters for Node Value Adjustment $a_1$, $a_2$}
\KwOut{Adjusted Node Value $E_{\text{Adjusted}}$
} 
\SetKwFunction{FRootwardPropagation}{RootwardPropagation}
\SetKwFunction{FLeafwardPropagation}{LeafwardPropagation}
\SetKwFunction{FAverage}{Average}
\SetKwProg{Fn}{Function}{:}{}

\Fn{\FRootwardPropagation{$v_i$}}{

$\vec{M} \gets \text{Empty list}~[~]$

\ForEach {$v_k \in \{v \in V \mid f(v)=v_i$\}}{

    $\vec{M}$ appends \FRootwardPropagation{$v_k$}
}

\If{$v_i$ is virtual node}{

$N \gets$ \FAverage{$\vec{M}$}

$E_{\text{Adjusted}}(v_i) \gets$ $N$

}

\Else{

$\vec{M}$ appends $E_{\text{Aligned}}(v_i)$

$N \gets$ \FAverage{$\vec{M}$}

}

\KwRet $N$

}

\Fn{\FLeafwardPropagation{$v_i$, $M$}}{

\If{$v_i$ is virtual node}{

$M \gets M\cdot E_{\text{Adjusted}}(v_i)$

\If{$v_i$ is first child node}{

$M \gets M\cdot a_2$

}

}

\Else{

$E_{\text{Adjusted}}(v_i) \gets E_{\text{Aligned}}(v_i) \cdot M^{a_1}$

}

\ForEach {$v_k \in \{v \in V \mid f(v)=v_i$\}}{

    \FLeafwardPropagation{$v_k$, $M$}
}

}

\FRootwardPropagation{$v_\text{root}$}

\FLeafwardPropagation{$v_\text{root}$, $1$}

\end{algorithm}

\subsection{Tree Compression Based on Node Value}
\label{TreeCompression}
After obtaining the global parse tree as well as the length and the value of each node, the selective prompt compression problem, as in Definition 3, can be then re-formulated as a tree pruning problem that considers the length and the value of the nodes. Formally, it can be formulated as follows.

\noindent\textbf{Definition 7. 
Parse Tree Pruning Problem
} 
Given a tree $T=(R,V,v_\text{root},f)$ for a specified prompt $R$ and a specified compression ratio $\tau$ where $0<\tau<1$, the task is to derive a compressed tree $T_{\text{cp}}=(R_{\text{cp}},V_{\text{cp}},v_\text{root,cp},f_{\text{cp}})$ for a selective compression $R_\text{cp}$. The compressed tree is the optimal solution to the following optimization problem
\begin{align}
\label{ValueLimit}
\max~&E(V_{\text{cp}})\\
\label{LengthLimit}
\text{s.t.}~&C(V_{\text{cp}})\leq \tau C(V),\\
\label{TreeLimit}
&f_{\text{cp}}:V_{\text{cp}}\setminus \{v_\text{root,cp}\} \mapsto V_{\text{cp}}.
\end{align}

The objective function (\ref{ValueLimit}) and constraint (\ref{LengthLimit}) are derived from the general selective prompt compression problem. The constraint (\ref{LengthLimit}), which corresponds to the constraint $C(R_{\text{cp}})\leq \tau C(R)$ in the general problem, ensures that the actual compression ratio $C(V_{\text{cp}})~/~C(V)$ does not exceed the given compression ratio $\tau$, 
thereby allowing precise control over the length of the compressed prompt. The objective function (\ref{ValueLimit}), which corresponds to objective function $\max E(R_{\text{cp}})$ in the general problem, aims to preserve the value of the nodes as much as possible. 

The constraint (\ref{TreeLimit}) guarantees that the retained nodes maintain the tree structure, a requirement derived from the property of the parse tree. This reflects the inherent dependencies within the parse tree, whereby the tokens on the child nodes are considered dependent on the tokens of the parent nodes. Consequently, if the compressed prompt includes tokens from the child nodes, the corresponding parent node tokens should also be preserved.

A recursive algorithm is then developed to derive the optimal solution. The pseudocode is provided in Algorithm \ref{RecursiveTreeCompression2}.
The output of the algorithm is a list, denoted by $Q$, which is defined as below: $Q = [Q_0, Q_1, Q_2, \dots]$, where each $Q_l$ represents the optimal solution limited to length $l$.
To be more specific, $Q_l=(Q_{l,\text{1}},Q_{l,\text{2}})=(E(V_{\text{cp},l}),V_{\text{cp},l})$, with $V_{\text{cp},l}$ being the node set of the optimal compressed tree limited to length $l$. Thus, the tree constructed by node set $Q_{\lfloor \tau C(V)\rfloor,\text{2}}$ is the optimal solution of the parse tree pruning problem in Definition 7. 
After that, the compressed prompt is obtained by concatenating the tokens corresponding to the aforementioned nodes in sequence. 
In Algorithm \ref{RecursiveTreeCompression2}, each call of function \texttt{CalculateSolution} recursively derives the optimal solution for the subtree rooted at $v_i$ from the optimal solutions of the subtrees rooted at all child nodes of $v_i$. This process is divided into two steps: merging the solutions from the child nodes using the function \texttt{MergeSolution} (Lines 11-13), and incorporating $v_i$ into the merged solution (Lines 14-18). Since both steps maintain the optimal state, the final result should also be optimal.

\begin{algorithm}
\caption{Recursive Tree Compression}
\label{RecursiveTreeCompression2}
\KwIn{Parse Tree $T=(R,V,v_\text{root},f)$, Length Function $C$, Adjusted Value $E$}
\KwOut{
List $Q=[Q_0,\cdots Q_l,\cdots]$ where $Q_l$ is the optimal solution limited to length $l$
}
\SetKwFunction{FMergeSolution}{MergeSolution}
\SetKwFunction{FCalculateSolution}{CalculateSolution}
\SetKwProg{Fn}{Function}{:}{}

\noindent\textbf{Definition} For a list $Q$, $C(Q)$ is defined as the length of $Q$, e.g. $C([Q_0,Q_1,\cdots ,Q_n])=n+1$

\Fn{\FMergeSolution{$Q$, $P$}}{

\ForEach{$0\leq l \leq (C(Q)+C(P)-2)$}{
$k \gets \underset{z}{\mathrm{argmax}}(Q_{z,1}+P_{l-z,1})$

$S_l \gets ((Q_{k,1}+P_{l-k,1}),(Q_{k,2}\cup P_{l-k,2}))$
}

$S \gets [S_0,\cdots ,S_{C(Q)+C(P)-2}]$

\KwRet $S$

}

\Fn{\FCalculateSolution{$v_i$}}{
$Q_0 \gets (0,\emptyset)$

$Q \gets [Q_0]$

\ForEach {$v_k \in \{v \in V \mid f(v)=v_i\}$}{
    $P \gets $ \FCalculateSolution{$v_k$}
    
    $Q \gets $ \FMergeSolution{$Q$, $P$}
}

\ForEach{$0\leq l \leq (C(Q)+C(v_i)-1)$}{

\If{$l<C(v_i)$}{
$S_l=(0,\emptyset)$
}

\Else{
$S_l=(E(v_j)+Q_{l-C(v_i),1},\{v_j\}\cup Q_{l-C(v_i),2})$
}

}

$S \gets [S_0,\cdots ,S_{C(Q)+C(P)-1}]$

\KwRet $S$
}

\FCalculateSolution{$v_\text{root}$}
\end{algorithm}

\subsection{Algorithm and Complexity Analysis of PartPrompt}
The complete procedure of the PartPrompt method is
presented in Algorithm \ref{PartPromptline}. 
For clarity, Section \ref{InformationEntropyApproximation} describes Line 3, where function \texttt{SmallModel} calculates the entropy of token by Equation (\ref{eq8}); 
Section \ref{GlobalParseTree} describes Line 4 and Line 6, where function \texttt{SentenceParser} builds the local parse tree and function \texttt{BuildGlobalTree} builds the hierarchical global tree; 
Section \ref{TokenAlignment} describes Line 5; 
Section \ref{InformationEntropyAdjustment} describes Line 7; 
and Section \ref{TreeCompression} describes Line 8. 
Moreover, function \texttt{PromptStructureParser} slices the original prompt into sentences, and function \texttt{TokenConcatenation} concatenates tokens into compressed prompts. 
The innovative recursive tree compression technique allows for the concurrent management of multiple compression ratios without additional computational costs. Users are thus able to compare the results of several compression ratios before selecting one.

\begin{algorithm*}
\caption{PartPrompt}
\label{PartPromptline}
\KwIn{Original Prompt $R$, Compression Ratio List $[\tau_1,\tau_2,\cdots,\tau_u]$, Node Value Adjustment Parameters $a_1$, $a_2$}
\KwOut{Compressed Prompts List of each Compression Ratio $[R_{cp,1},R_{cp,2},\cdots,R_{cp,u}]$} 

\SetKwFunction{FPromptStructureParse}{PromptStructureParser}
\SetKwFunction{FSmallModel}{SmallModel}
\SetKwFunction{FParser}{SentenceParser}
\SetKwFunction{FTokenAlignment}{TokenAlignment}
\SetKwFunction{FBuildGlobalTree}{BuildGlobalTree}
\SetKwFunction{FNodeValueAdjustment}{NodeValueAdjustment}
\SetKwFunction{FRecursiveTreeCompression}{RecursiveTreeCompression}
\SetKwFunction{FTokenAssemble}{TokenConcatenation}

Sentence List $[R_1,R_2,\cdots,R_m]$, Structure Label List $[g_1,g_2,\cdots,g_m]$ $\gets$ \FPromptStructureParse{$R$}

\ForEach {Sentence $R_j$ in Sentence List}{
Token List $[s_1,s_2,\cdots,s_{l_j}]_j$, Entropy List $[e_1,e_2,\cdots,e_{l_j}]_j$ $\gets$ \FSmallModel{$R_j$}

Node List $[r_1,r_2,\cdots,r_{n_j}]_j$, Edge List $[(r_{w_1},r_{w_2}),(r_{w_2},r_{w_3}),\cdots,(r_{w_{h_j}},r_{w_{n_j}})]_j$ $\gets$ \FParser{$R_j$}

Aligned Entropy List $[\tilde{e}_1,\tilde{e}_2,\cdots,\tilde{e}_{n_j}]_j$, Length List $[C_1,C_2,\cdots,C_{n_j}]_j \gets$ \FTokenAlignment{Token List, Node List, Entropy List}

}

Global tree $T \gets$ \FBuildGlobalTree{Edge Lists, Structure Label List, Node Lists}

Adjusted Node Value Lists $[\hat{e}_1,\hat{e}_2,\cdots,\hat{e}_n] \gets$ \FNodeValueAdjustment{$T$, Aligned Entropy Lists, $a_1$, $a_2$}

Solution $Q$ $\gets$ \FRecursiveTreeCompression{$T$, Length Lists, Adjusted Node Value Lists}

Compressed Prompts List $[R_{cp,1},R_{cp,2},\cdots,R_{cp,u}]$ $\gets$ \FTokenAssemble{Solution, Compression Ratio List}

\end{algorithm*}
To analyze the computational complexity of each component within the process described in Algorithm \ref{PartPromptline}, let us consider a prompt composed of $m$ sentences, with each sentence containing $n$ tokens. The computational complexity of Algorithm \ref{PartPromptline} is broken down as follows: Line 1, Line 6 requires $\mathcal{O}(m)$; Line 3 requires $\mathcal{O}(Kmn^2)$, where $K$ is related to the size of language model; Line 4 requires $\mathcal{O}(mn^3)$; Line5, Line 7, and Line 9 requires $\mathcal{O}(mn)$; Line 8 requires $\mathcal{O}(m^2n^2)$. In comparison, the computational complexity of Selective-Context \cite{li2023compressingcontextenhanceinference} is $\mathcal{O}(Km^2n^2)$, while LLMLingua \cite{jiang2023llmlingua} requires 
$\mathcal{O}(Kmn^2(1+(m-1)\tau+(m-2)\tau^2))$, with $\tau$ representing the compression ratio, and LLMLingua2\cite{pan2024llmlingua} requries $\mathcal{O}(Km^2n^2)$. Further technical details are elaborated in the Appendix A. 
Note that, under the typical conditions, $m$ and $n$ range around $20\sim50$, $\tau$ lies between $0.2\sim0.5$, and $K \gg m,n$. Therefore, the overall computational complexity of the PartPrompt method is $\mathcal{O}(Kmn^2)$, resulting in the following rrelationship of computational demands among these four methods as $\mathcal{O}(\text{PartPrompt})<\mathcal{O}(\text{LLMLingua})<\mathcal{O}(\text{Selective-Context})\approx\mathcal{O}(\text{LLMLingua2})$.

\subsection{Theoretical Analysis}
\label{TheoreticalAnalysis}

This section elaborates the theoretical analysis of information entropy approximation and the theoretical connections of PartPrompt to the two prominent selective prompt compression methods: Selective-Context \cite{li2023compressingcontextenhanceinference} and LLMLingua \cite{jiang2023llmlingua}.

In the information entropy approximation module, the term $r_{<j}$ is excluded when calculating $E(r_{j,i})$. This removed term is same for all tokens $r_{j,1}, r_{j,2}, \cdots ,r_{j,n_j}$ within the same sentence, indicating a roughly consistent approximation error across these tokens. However, for tokens across different sentences, those with a larger $j$ have more $r_{<j}$ omitted, leading to a greater neglect of information and a potential overestimation of $E(r_{j,i})$. Considering that the objective Function (\ref{ValueLimit}) of tree compression aims to preserve nodes with larger value, the approximation error leads to a bias towards retaining tokens with a lager $j$. Therefore, this inter-sentence error need to be managed efficiently, or may results in performance loss.

Next, we present the theoretical connections of PartPrompt to the prominent methods. Selective-Context retains tokens with higher information entropy and employs a parse-based tokenizer to ensure token completeness \cite{li2023compressingcontextenhanceinference}. 
Accordingly, Selective-Context can be reformulated within our parse tree compression framework as follows.
For a given prompt $R$, a tree $T=(R,V,v_\text{root},f)$ is constructed, where $v_\text{root}$ is the only virtual node, and all actual nodes $v_i$ are its child nodes with $f(v_i)=v_\text{root}$.
The tree pruning problem in Definition 7 is then solved with $E(v_i)$ calculated by Equation (\ref{EntropyComplete}).
Therefore, Selective-Context can be considered as a simplified version of PartPrompt that flattens the global parse tree and omits the approximation of information entropy.

LLMLingua, another representative selective prompt compression method, consists of the budget controller, iterative token-level prompt compression, and distribution alignment \cite{jiang2023llmlingua}.
The budget controller divides the prompt into paragraphs, sorting them by information entropy and preserving those with higher entropy.  In PartPrompt, the node value adjustment algorithm for actual nodes is defined as $E_{\text{Adjusted}}(v_i) \gets E_{\text{Aligned}}(v_i) \cdot M^{a_1}$ (Line 18 of Algorithm \ref{AdjustmentAlgorithm}). Here $M$ is derived from the root-ward propagation and convergence of the information entropy of the corresponding segemts. 
In light of the fact that $M > 1$ is typically the case, 
there exists a sufficiently large $a_1$ such that for any $i,j, M_i>M_j$ leads to $E_{\text{Adjusted}}(v_i)>E_{\text{Adjusted}}(v_j)$.
In this case, parse tree compression is equivalent to retaining the part with the larger value of $M$. 
Therefore, the budget controller of LLMLingua can be considered as a variant of parse tree compression, which 
involves flattening the virtual part of the global parse tree 
and setting parameter $a_1\gg 1$ in the node value adjustment module of PartPrompt.

Considering $r_{j,i}$ as the $i$-th token of the $j$-th sentence, the iterative token-level prompt compression of LLMLingua employs Equation (\ref{EntropyLLMLingua}) to calculate the information entropy. 
This approach could be regarded as an intermediate stage between the initial version employed by Selective-Context, i.e., Equation (\ref{EntropyComplete}), and the local approximation version of PartPrompt, i.e., Equation (\ref{EntropyApprox}).
\begin{equation}
\label{EntropyLLMLingua}
E(r_{j,i})=-\log~p_{\text{LM}}(~r_{j,i}~|~r_{j,<i},r_{<j,\text{retained}}).
\end{equation}
\begin{equation}
\label{EntropyComplete}
E(r_{j,i})=-\log~p_{\text{LM}}(~r_{j,i}~|~r_{j,<i},r_{<j}).
\end{equation}
\begin{equation}
\label{EntropyApprox}
E(r_{j,i})=-\log~p_{\text{LM}}(~r_{j,i}~|~r_{j,<i}).
\end{equation}

It might be worth mentioning that the approximation from Equation (\ref{EntropyComplete}) to Equation (\ref{EntropyApprox}) does not lead to a significant performance loss, when the approximation error is effectively managed as demonstrated by the empirical study in Section \ref{AblationStudy}. On the other hand, the approximation significantly reduces the computational cost, since the entropy is computed within each sentence rather than the whole prompt.

\section{Experimental Settings}
\subsection{Datasets}

Four representative datasets are employed to evaluate the proposed method. Concretely, BBCnews dataset is collected from articles on the BBC News website \cite{BBC2024BBCnews}, and arXiv dataset is collected from papers on the arXiv preprint platform \cite{Cornell2024arXiv}. 
Following Selective-Context \cite{li2023compressingcontextenhanceinference} and LLMLingua \cite{jiang2023llmlingua}, these two datasets are used for testing contextual understanding prompts with the task of summarizing articles.
To ensure that the model has not seen these data during training, we re-crawl new data for BBC News and arXiv, with all the release dates of these data occurring after Jan 1, 2024. Due to the length of arXiv dataset, which exceeds the input length that Selective-Context and LLMLingua can handle \cite{li2023compressingcontextenhanceinference,jiang2023llmlingua}, we truncate the first 3000 tokens of the main text, referring as truncate arXiv. The original arXiv dataset is also used to verify the applicability of our method for texts exceeding the input limit of compared methods.
The PeopleDaily dataset is a dataset in Chinese constructed by crawling articles from the People's Daily newspaper\cite{PeoplesDaily2024PeoplesDaily}. CodeNet is a code dataset. We clean a subset of CodeNet\cite{puri2021codenet} to create the version used in this work. These two datasets are employed to evaluate the performance on non-English and non-natural language prompt.

There are four additional important datasets, which are tested in appendix. 
HotpotQA \cite{yang-etal-2018-hotpotqa} is a multi-hop question answering dataset, which is designed to test the performance for question answering and multi-hop reasoning prompts. We use the same setting as in  \cite{li2023leveraging} and select 500 questions for testing. GSM8K \cite{cobbe2021trainingverifierssolvemath} is a classic mathematical reasoning dataset. Following \cite{jiang2023llmlingua}, we adopt the chain-of-thought prompt provided by \cite{fu2023chainofthoughthubcontinuouseffort} to evaluate the performance for in-context learning and chain-of-thought prompts.
RULER \cite{hsieh2024ruler} and LongBecnh \cite{bai2023longbench} are two important long prompt datasets. 

\subsection{Compared Methods for Prompt Compression}
As our method belongs to selective prompt compression, we compare it to the state-of-the-art selective prompt compression methods. Selective-context \cite{li2023compressingcontextenhanceinference} is the first to propose selective prompt compression method. It uses a smaller LLM to calculate the information entropy of tokens and removes words with lower information entropy. The question text is fully preserved for better performance. LLMLingua \cite{jiang2023llmlingua} further proposes the model distribution alignment, budget controller and token-level iterative compression algorithm based on Selecting-context. LLMLingua receives significant improvements in context-understanding prompts, and makes selective compression method viable for chain-of-thought prompts. 
LLMLingua2 \cite{pan2024llmlingua}, which achieves word selective compression by distilling the compression results of GPT4 to a small model, is a completely new line approach and represents the current state-of-the-art level of prompt compression.
LongLLMLingua \cite{jiang2023longllmlingua} is also an important compression method, but unlike the above methods, it belongs to the question-aware approach.

In addition to selective compression methods, our method is also compared with the generative compression method. Qwen2-72B \cite{yang2024qwen2technicalreport}, the up-to-date generative LLM trained by Alibaba, is exploited to directly compress the original prompt. To be more specific, we feed the original prompt into Qwen2-72B together with the prompt of asking LLM for compression\footnote{The prompt of asking LLM for compression is as follows: "Eliminate repetitive elements and present the text concisely, ensuring that key details and logical processes are retained.“, which comes from \cite{jiang2023llmlingua}.} to generate the compressed prompt. To further highlight the advantage of our method, we compare it with the generative compression method over the four normal datasets, despite using LLMs like Qwen2-72B for compression requires huge computational resources. Besides, we additionally conduct experiments for a very long prompt scenario that most selective compression methods are unable to manage. 

\subsection{Target Models for Inference}
The prompt is employed to assist the inference and generation of target LLMs. In fact, the same prompt for different target LLMs would have different performances. To verify the effectiveness of our method across various target LLMs, we conduct experiments for the following LLMs: Mixtral-8x7B \cite{jiang2024mixtralexperts}, Llama3-70B \cite{meta2024llama3}, and Qwen2-72B \cite{yang2024qwen2technicalreport}. Among them, Mixtral-8x7B is a famous mix-of-expert LLM performing well on various tasks, and is set as the default model unless specified otherwise. Llama3-70B is a recent high-performance LLM trained by Meta; Qwen2-72B is the up-to-date LLM trained by Alibaba and receives the higher performance than Llama3-70B according to the HELM Leaderboard\footnote{HELM Leaderboard: https://crfm.stanford.edu/helm/mmlu/v1.6.0}.

\subsection{Evaluation}
Following \cite{li2023compressingcontextenhanceinference,jiang2023llmlingua}, we take BLEU \cite{papineni2002bleu}, Rouge (including Rouge1, Rouge2, RougeL) \cite{lin2004rouge}, and BERTScore (specifically BS-F1) \cite{zhang2019bertscore} as the metric for evaluation on BBCnews, (truncate) arXiv, PeoPleDaily and CodeNet datasets.
BLEU is a composite metric of four metrics: 1-gram, 2-gram, 3-gram, and 4-gram, and is often used in the natural language processing. Rouge is classic linguistic metrics based on rule matching, while BERTScore calculates the similarity of the embeddings from BERT.
The ground-truth answer is set to the output of the uncompressed prompt for the above four datasets. The evaluation for another four datasets can be found in appendix.

\subsection{Other Settings}\label{sec:implement}
Apart from above settings, other experimental settings are clarified as follows. The information entropy is calculated using Llama2-7B~\cite{touvron2023llama}, and the local parse tree is analyzed using the Stanford CoreNLP \cite{manning2014stanford}. 
When adjusting the value in parse trees, the related hyper-parameters are set as follows: for BBCnews, $a_1=4$, $a_2=100$; for truncate arXiv, $a_1=3$, $a_2=100$; 
for PeopleDaily, $a_1=3$, $a_2=25$; for CodeNet, $a_1=1$, $a_2=5$. 
And the typical range of them are $0\leq a_1\leq 5$, $1\leq a_2\leq 1000$.
The maximum token length for target LLMs is set to 300 during inference for all experiments.
For the pretrained language models used in this work, Mixtral-8x7B and Llama3-70B are accessed through HuggingFace Pro API\footnote{Mixtral-8x7B and Llama3-70B: https://huggingface.co/blog/inference-pro}; Qwen2-72B is accessed via Aliyun or can be downloaded from HuggingFace\footnote{Qwen2-72B: https://huggingface.co/Qwen/Qwen2-72B-Instruct}; Llama2-7B is downloaded from HuggingFace\footnote{Llama2-7B: https://huggingface.co/meta-llama/Llama-2-7b} and run on an NVIDIA GeForce RTX 3090 GPU with 24G memory. More settings, e.g, using DeepSeek-LLM-7B\cite{bi2024deepseek} to calculate information entropy, can be found in appendix.

\begin{table*}[htbp]
\centering
\caption{
The performance of PartPrompt in comparison with baseline methods on multiple datasets, compression ratios, and metrics. The \textbf{best performance} among three selective methods is in bold. PartPrompt achieves the best performance in almost all scenarios. The baseline of LLM Generation is only as a reference, as using generative LLMs for prompt compression is not practical.
}
\renewcommand{\arraystretch}{1.15}
\setlength{\aboverulesep}{0pt}
\setlength{\belowrulesep}{0pt}
\scalebox{0.96}{
\begin{tabular}{>{\centering}m{2cm}
>{\centering}m{0.7cm}
>{\centering}m{0.8cm}
>{\centering}m{0.8cm}
>{\centering}m{0.8cm}
>{\centering}m{0.75cm}
>{\centering}m{0.75cm}
>{\centering}m{0.6cm}|
>{\centering}m{0.7cm}
>{\centering}m{0.8cm}
>{\centering}m{0.8cm}
>{\centering}m{0.8cm}
>{\centering\arraybackslash}m{0.75cm}
>{\centering}m{0.75cm}
>{\centering\arraybackslash}m{0.6cm}
}
\toprule
\multirow{2}{*}{methods}&\multicolumn{7}{c|}{BBCnews}&\multicolumn{7}{c}{truncate arXiv}\\
&BLEU&Rouge1&Rouge2&RougeL&BS-F1&Tokens&$1/\tau$&BLEU&Rouge1&Rouge2&RougeL&BS-F1&Tokens&$1/\tau$\\
\hline
LLM Generation& 11.31  & 42.32  & 18.60  & 39.22  & 69.17  & 261.3  & 3.29  & 8.77  & 39.18  & 16.00  & 36.22  & 68.21  & 643.8  & 4.65  \\
\hline
\multicolumn{15}{c}{20\% tokens constraint} \\
\hline
Selective-Context & 3.58  & 31.63  & 9.22  & 28.73  & 60.64  & 188.6  & 4.56  & 5.28  & 34.86  & 12.25  & 31.48  & 65.10  & 656.6  & 4.56  \\
LLMLingua & 9.56  & 35.09  & 15.28  & 32.89  & 61.97  & 206.0  & 4.18  & 7.35  & 36.10  & 14.15  & 33.18  & 65.03  & 576.7  & 5.19  \\
LongLLMLingua & 9.22  & 33.49  & 14.60  & 31.42  & 60.66  & 161.0  & 5.34  & 11.22  & 39.49  & 18.47  & 36.77  & 67.13  & 599.6  & 5.00  \\
LLMLingua2 & 5.65  & 36.22  & 11.95  & 32.82  & 63.52  & 172.6  & 4.99  & 5.75  & 35.96  & 12.39  & 32.63  & 65.76  & 615.1  & 4.87  \\
\textbf{PartPrompt} & \textbf{13.69 } & \textbf{43.54 } & \textbf{22.87 } & \textbf{41.05 } & \textbf{67.58 } & 173.7  & 4.96  & \textbf{11.73 } & \textbf{39.83 } & \textbf{18.74 } & \textbf{37.01 } & \textbf{67.32 } & 595.8  & 5.03  \\
\hline
\multicolumn{15}{c}{30\% tokens constraint} \\
\hline
Selective-Context & 6.68  & 37.49  & 13.54  & 34.15  & 64.32  & 294.2  & 2.93  & 7.65  & 38.14  & 14.92  & 34.65  & 66.64  & 1048.1  & 2.86  \\
LLMLingua & 10.22  & 38.55  & 16.66  & 35.88  & 64.64  & 278.2  & 3.09  & 10.39  & 39.23  & 17.47  & 36.26  & 66.95  & 867.9  & 3.45  \\
LongLLMLingua & 10.11  & 36.97  & 15.96  & 34.48  & 63.50  & 257.8  & 3.34  & 14.62  & 42.61  & 22.13  & 39.86  & 68.73  & 894.9  & 3.35  \\
LLMLingua2 & 9.45  & 41.86  & 16.71  & 38.39  & 66.88  & 258.1  & 3.34  & 7.86  & 38.86  & 14.91  & 35.63  & 67.21  & 898.4  & 3.33  \\
\textbf{PartPrompt} & \textbf{18.38 } & \textbf{48.57 } & \textbf{28.39 } & \textbf{46.21 } & \textbf{70.20 } & 257.7  & 3.34  & \textbf{15.35 } & \textbf{43.12 } & \textbf{22.50 } & \textbf{40.26 } & \textbf{68.95 } & 839.4  & 3.57  \\
\hline
\multicolumn{15}{c}{50\% tokens constraint} \\
\hline
Selective-Context & 14.26  & 46.74  & 22.63  & 43.41  & 69.53  & 481.0  & 1.79  & 13.63  & 44.56  & 21.78  & 41.32  & 69.95  & 1763.7  & 1.70  \\
LLMLingua & 15.71  & 45.98  & 23.21  & 43.17  & 69.52  & 435.8  & 1.97  & 17.26  & 45.85  & 25.00  & 43.08  & 70.46  & 1445.9  & 2.07  \\
LongLLMLingua & 14.65  & 44.31  & 21.27  & 41.33  & 68.22  & 432.7  & 1.99  & 18.79  & 46.71  & 26.48  & 43.82  & 70.92  & 1497.4  & 2.00  \\
LLMLingua2 & 17.65  & 50.12  & 26.06  & 46.89  & 71.39  & 430.9  & 2.00  & 14.07  & 44.95  & 21.92  & 42.03  & 70.16  & 1500.9  & 2.00  \\
\textbf{PartPrompt} & \textbf{28.28 } & \textbf{56.48 } & \textbf{38.30 } & \textbf{54.29 } & \textbf{74.57 } & 424.9  & 2.03  & \textbf{19.50 } & \textbf{46.78 } & \textbf{26.87 } & \textbf{44.26 } & \textbf{70.93 } & 1405.8  & 2.13  \\
\hline
\hline
\multirow{2}{*}{methods}&\multicolumn{7}{c|}{PeopleDaily}&\multicolumn{7}{c}{CodeNet}\\
&BLEU&Rouge1&Rouge2&RougeL&BS-F1&Tokens&$1/\tau$&BLEU&Rouge1&Rouge2&RougeL&BS-F1&Tokens&$1/\tau$\\
\hline
LLM Generation& 10.11  & 36.57  & 15.75  & 33.11  & 67.77  & 795.5  & 2.97  & 7.78  & 36.45  & 14.81  & 33.69  & 63.27  & 820.7  & 1.85  \\
\hline
\multicolumn{15}{c}{20\% tokens constraint} \\
\hline
Selective-Context & 3.06  & 26.43  & 8.10  & 23.83  & 61.06  & 477.2  & 4.94  & 4.32  & 29.68  & 10.14  & 27.60  & 58.50  & 341.4  & 4.45  \\
LLMLingua & 4.20  & 28.21  & 9.67  & 25.55  & 62.28  & 471.9  & 5.00  & 5.17  & 31.16  & 11.51  & 29.06  & 58.99  & 303.1  & 5.01  \\
LongLLMLingua & 6.88  & 32.03  & 12.84  & 29.11  & 64.32  & 473.1  & 4.99  & 4.39  & 30.11  & 10.42  & 27.99  & 58.33  & 303.8  & 5.00  \\
LLMLingua2 & 3.21  & 27.57  & 8.73  & 24.74  & 61.83  & 504.7  & 4.67  & 2.77  & 27.48  & 8.39  & 25.42  & 57.32  & 310.3  & 4.90  \\
\textbf{PartPrompt} & \textbf{12.69 } & \textbf{38.87 } & \textbf{20.11 } & \textbf{36.25 } & \textbf{67.98 } & 471.9  & 5.00  & \textbf{6.48 } & \textbf{34.00 } & \textbf{13.11 } & \textbf{31.59 } & \textbf{60.86 } & 312.5  & 4.86  \\
\hline
\multicolumn{15}{c}{30\% tokens constraint} \\
\hline
Selective-Context & 3.74  & 28.49  & 9.08  & 25.25  & 62.45  & 697.0  & 3.38  & 5.60  & 31.62  & 11.78  & 29.44  & 59.72  & 461.4  & 3.29  \\
LLMLingua & 5.81  & 30.92  & 11.56  & 28.09  & 64.33  & 707.3  & 3.33  & 6.62  & 33.05  & 13.24  & 30.84  & 60.02  & 456.9  & 3.33  \\
LongLLMLingua & 7.93  & 33.37  & 13.85  & 30.49  & 65.41  & 703.8  & 3.35  & 5.43  & 31.01  & 11.42  & 28.82  & 58.65  & 456.0  & 3.33  \\
LLMLingua2 & 4.06  & 29.50  & 10.08  & 26.57  & 63.25  & 708.0  & 3.33  & 4.23  & 29.99  & 10.46  & 27.80  & 59.06  & 471.4  & 3.22  \\
\textbf{PartPrompt} & \textbf{16.39 } & \textbf{42.29 } & \textbf{23.34 } & \textbf{39.48 } & \textbf{69.97 } & 710.0  & 3.32  & \textbf{8.29 } & \textbf{36.70 } & \textbf{15.50 } & \textbf{34.30 } & \textbf{62.33 } & 470.2  & 3.23  \\
\hline
\multicolumn{15}{c}{50\% tokens constraint} \\
\hline
Selective-Context & 5.07  & 31.02  & 10.86  & 27.85  & 64.09  & 1191.1  & 1.98  & 7.30  & 34.37  & 13.95  & 31.91  & 61.18  & 758.4  & 2.00  \\
LLMLingua & 10.34  & 36.84  & 16.44  & 33.71  & 67.85  & 1180.0  & 2.00  & 8.39  & 35.60  & 15.38  & 33.21  & 61.71  & 758.5  & 2.00  \\
LongLLMLingua & 11.85  & 37.75  & 17.90  & 34.61  & 67.89  & 1181.5  & 2.00  & 7.03  & 34.06  & 13.93  & 31.80  & 60.90  & 757.2  & 2.01  \\
LLMLingua2 & 5.80  & 31.95  & 11.75  & 28.65  & 64.76  & 1151.3  & 2.05  & 5.65  & 32.35  & 12.23  & 30.07  & 60.03  & 770.3  & 1.97  \\
\textbf{PartPrompt} & \textbf{20.68 } & \textbf{46.66 } & \textbf{27.56 } & \textbf{43.82 } & \textbf{72.47 } & 1179.1  & 2.00  & \textbf{12.05 } & \textbf{41.95 } & \textbf{20.04 } & \textbf{39.40 } & \textbf{65.59 } & 778.0  & 1.95\\

\hline
\end{tabular}
}
\scalebox{0.96}{
\begin{tabular}{>{\centering}m{2cm}
>{\centering}m{0.7cm}
>{\centering}m{0.8cm}
>{\centering}m{0.8cm}
>{\centering}m{0.8cm}
>{\centering}m{0.75cm}
>{\centering}m{0.75cm}
>{\centering}m{0.6cm}
>{\centering}m{0.7cm}
>{\centering}m{0.8cm}
>{\centering}m{0.8cm}
>{\centering}m{0.8cm}
>{\centering\arraybackslash}m{0.75cm}
>{\centering}m{0.75cm}
>{\centering\arraybackslash}m{0.6cm}
}
\hline
&&&&&&&&&&&&&&
\end{tabular}
}
\label{SOTAexperiment}
\end{table*}

\section{Experiments}\label{sec:exp}
Extensive experiments are conducted to demonstrate the effectiveness and superiority of PartPrompt (the proposed method) regarding various prompt compression methods, datasets, metrics, compression ratios, and scenarios. Section \ref{Main Experiments} presents a thorough comparison between PartPrompt and other compression methods to confirm the superiority of the proposed method. Considering the effect of compression ratios and target LLMs for inference, PartPrompt is further compared with other methods for various compression ratios in Section \ref{VariousCompressionRatiosStudy} and different target LLMs in Section \ref{targetmodelstudy}. Section \ref{AblationStudy} provides a comprehensive ablation study to verify the positive role of each component in PartPrompt. Section \ref{InputLimitation} investigates the potential of PartPrompt under the extreme long prompt scenario. In Section \ref{coherencestudy} and \ref{Visualizationstudy}, the compressed prompt is directly compared with uncompressed one; accordingly more abundant metrics and an intuitive example are employed to show the advantages of PartPrompt in terms of text similarity and coherence maintenance.

\subsection{Comparative Study: Main Experiments}
\label{Main Experiments}
Comprehensive experiments are conducted in this section to fully compare the proposed method with other selective compression methods. Four diverse datasets and multiple metrics are employed to evaluate the performance. For clarity, we choose three compression ratios for analysis: 20\%, 30\%, and 50\%. 
As the actual compression ratio is not under the control of some methods, 
the average length of the compressed prompt (named as “tokens”) and inverse compression ratio ($1/\tau$) are also listed to show the actual compression result. The LLM Generation is presented with a single compression ratio, since there is no compression ratio to set. 

The experimental results are shown in Table \ref{SOTAexperiment}. 
Overall, it is evident that PartPrompt achieves significant improvements across all datasets, compression ratios, and metrics, reaching state-of-the-art (SOTA) performance. Specifically, PartPrompt substantially outperforms the baseline methods in terms of BLEU and ROUGE scores. Notable gains are also observed on BERT scores with PartPrompt. 
In addition, the LLMLingua2 method, being the second generation of the LLMLingua series, has not made an improvement compared to LLMLingua. This is because LLMLingua2 is not, in the conventional sense, an upgraded version of LLMLingua; rather, it represents an entirely novel approach and is good at other scenarios.

Apart from comparing to selective compression methods, PartPrompt is further compared to the generative compression method. Qwen2-72B is employed for generating compressed prompts, which is denoted as LLM Generation in Table \ref{SOTAexperiment}. Note that, using an LLM for compression needs a significant amount of computational resources, which is contrary to the original goal of prompt compression and therefore impractical. Despite that, LLM Generation may still serve as a reference method.
These results show that PartPrompt outperforms LLM Generation on all four datasets, which not only proves the advantage of PartPrompt but also validates the rationality of selective compression methods.

\begin{figure*}[htbp]
    \centering
    \includegraphics[width=0.91\textwidth]{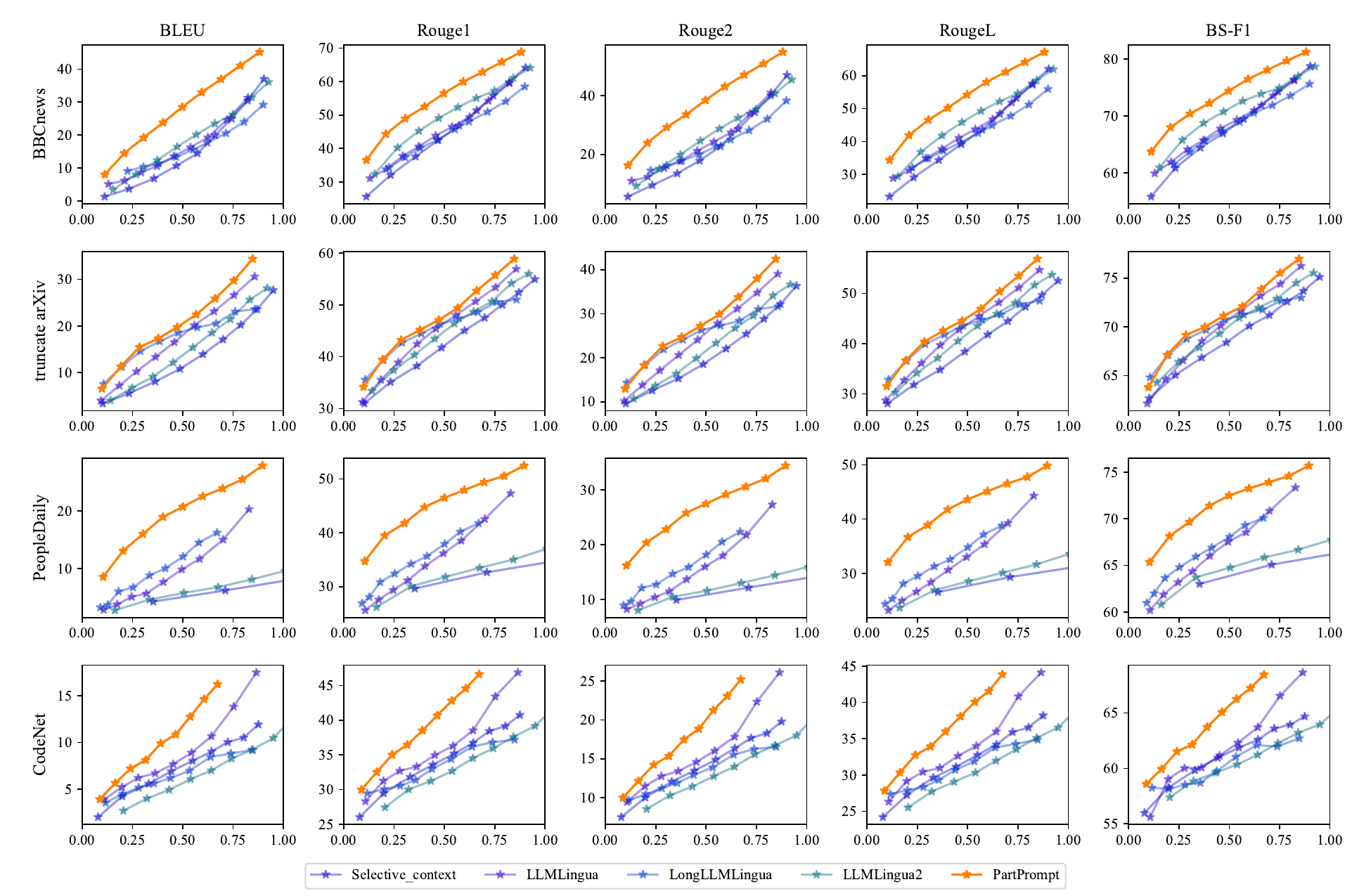}
    \caption{
    The performance of selective prompt compression methods over various compression ratios. The horizontal axis of each subplot depicts the ratios.
    }
    \label{3in1}
\end{figure*}

\subsection{Comparative Study: Under Various Compression Ratios}
\label{VariousCompressionRatiosStudy}
The compression ratio is a vital aspect in the prompt compression problem. To further investigate the effectiveness of these methods across a range of compression ratios, we carry out the experiments with nine different compression ratios, i.e., $\tau= 0.1, 0.2, 0.3, 0.4, 0.5, 0.6, 0.7, 0.8, 0.9$. 
The results for BBCnews, truncate arXiv, PeopleDaily and CodeNet are plotted in Figure \ref{3in1}.
The horizontal axis of each subplot depicts the compression ratio $\tau$ and the vertical axis corresponds to the metric score.

It can be observed that PartPrompt achieves considerable improvements for almost all compression ratios on these datasets, which again proves the superiority of the proposed method. 
Notably, the performance curve of PartPrompt displays a more pronounced convexity towards the upper left. This indicates that as the compression ratio decreases (i.e., the shorter compressed prompt), PartPrompt can better prioritize the deletion of less important parts, thereby slowing down the performance decline.
It is worth mentioning the low to medium compression ratios are more challenging cases than the high compression ratios. And comparing to other methods, the proposed PartPrompt tends to receive more performance gains under the low and medium compression ratios in BBCnews, truncate arXiv and PeopleDaily datasets. 
Furthermore, both Selective-Context and LLMLingua2 methods exhibit poor adaptability to PeopleDaily, the Chinese dataset, producing outputs with substantial unreadable texts. It leads to an actual compression ratio that is significantly greater than the target ratio and a significant decline in performance.

An interesting observation is the similarity in curves between BLEU and ROUGE-2, and between ROUGE-1 and ROUGE-L.
This pattern alignment can be attributed to their underlying computational principles. Specifically, both BLEU and ROUGE-2 rely on contiguous n-gram matching. Conversely, ROUGE-1 and ROUGE-L incorporate non-contiguous word matches and exhibit greater similarity to each other. 
The consistency of the underlying principles and results in turn renders the reliability of our experimental results. 

\subsection{Comparative Study: Using Different LLMs for Inference}
\label{targetmodelstudy}

The Prompt is designed to be used with an LLM, and the same prompt with different LLMs would have different performances. Consequently, it is necessary to evaluate the performance for different LLMs. In addition to Mixtral-8x7B as the default LLM, two additional LLMs are also employed in this section: Llama3-70B and Qwen2-72B. The experiments are conducted on BBCnews, and the ground truth is set as the output of the corresponding LLM given the uncompressed prompt. The results are presented in Table \ref{LLMBase}.

First, we observe that PartPrompt not only consistently outperforms baselines on Mixtral-8x7B but also consistently achieves better performances on Llama3-70B and Qwen2-72B. The observation substantiates the superiority and robustness of PartPrompt across different LLMs during inference. Second, it is worth mentioning that Selective-Context and LLMLingua yield quite low performances on Llama3-70B, which is caused by a large number of empty responses from Llama3-70B, a phenomenon not observed while feeding the compressed prompts given by PartPrompt to Llama3-70B.

\begin{table}[htbp]
  \centering
  \caption{The performance of selective prompt compression methods by feeding the compressed prompt to different LLMs.}
    \begin{tabular}{>{\centering\arraybackslash}m{2cm}
    >{\centering\arraybackslash}m{0.7cm}
    >{\centering\arraybackslash}m{0.85cm}
    >{\centering\arraybackslash}m{0.85cm}
    >{\centering\arraybackslash}m{0.85cm}
    >{\centering\arraybackslash}m{0.75cm}
}
    \toprule
    methods & BLEU & Rouge1 & Rouge2 & RougeL & BS-F1 \\
    \midrule
    \multicolumn{6}{c}{\textbf{Llama3-70B}} \\
    \midrule
    \multicolumn{6}{c}{20\% token constraint} \\
    Selective-Context & 1.74  & 12.04  & 3.85  & 11.01  & 22.16  \\
    LLMLingua & 2.64  & 6.75  & 3.50  & 6.31  & 11.45  \\
    LongLLMLingua & 3.05  & 26.23  & 8.02  & 24.40  & 47.02  \\
    LLMLingua2 & 4.87  & 30.94  & 11.17  & 28.50  & 51.92  \\
    \textbf{PartPrompt} & \textbf{13.64 } & \textbf{41.71 } & \textbf{22.07 } & \textbf{39.35 } & \textbf{65.88 } \\
    \midrule
    \multicolumn{6}{c}{30\% token constraint} \\
    Selective-Context & 1.88  & 9.11  & 3.48  & 8.26  & 15.09  \\
    LLMLingua & 4.06  & 13.79  & 6.07  & 12.82  & 23.77  \\
    LongLLMLingua & 3.66  & 29.71  & 9.47  & 27.68  & 50.57  \\
    LLMLingua2 & 8.23  & 35.92  & 15.96  & 33.50  & 54.40  \\
    \textbf{PartPrompt} & \textbf{18.26 } & \textbf{45.52 } & \textbf{26.58 } & \textbf{43.09 } & \textbf{67.04 } \\
    \midrule
    \multicolumn{6}{c}{50\% token constraint} \\
    Selective-Context & 1.28  & 4.06  & 1.97  & 3.79  & 6.16  \\
    LLMLingua & 8.19  & 21.25  & 11.21  & 19.99  & 31.92  \\
    LongLLMLingua & 5.52  & 35.05  & 12.83  & 32.55  & 55.19  \\
    LLMLingua2 & 13.60  & 43.29  & 24.25  & 40.84  & 57.80  \\
    \textbf{PartPrompt} & \textbf{25.87 } & \textbf{51.80 } & \textbf{33.67 } & \textbf{49.55 } & \textbf{70.87 } \\
    \midrule
    \midrule
    \multicolumn{6}{c}{\textbf{Qwen2-72B}} \\
    \midrule
    \multicolumn{6}{c}{20\% token constraint} \\
    Selective-Context & 2.23  & 27.34  & 6.54  & 24.16  & 57.95  \\
    LLMLingua & 6.40  & 28.99  & 10.57  & 26.47  & 58.43  \\
    LongLLMLingua & 3.63  & 25.77  & 7.21  & 23.49  & 55.69  \\
    LLMLingua2 & 5.55  & 35.37  & 10.98  & 31.23  & 63.75  \\
    \textbf{PartPrompt} & \textbf{9.27 } & \textbf{37.29 } & \textbf{15.60 } & \textbf{34.13 } & \textbf{65.56 } \\
    \midrule
    \multicolumn{6}{c}{30\% token constraint} \\
    Selective-Context & 4.97  & 33.14  & 10.21  & 29.66  & 62.68  \\
    LLMLingua & 6.91  & 32.00  & 11.78  & 29.12  & 61.10  \\
    LongLLMLingua & 4.11  & 28.35  & 8.44  & 25.77  & 58.38  \\
    LLMLingua2 & 9.64  & 40.98  & 15.81  & 36.76  & 67.86  \\
    \textbf{PartPrompt} & \textbf{13.51 } & \textbf{41.84 } & \textbf{20.03 } & \textbf{38.55 } & \textbf{68.70 } \\
    \midrule
    \multicolumn{6}{c}{50\% token constraint} \\
    Selective-Context & 11.72  & 42.88  & 18.55  & 39.08  & 69.01  \\
    LLMLingua & 11.29  & 40.06  & 17.35  & 36.83  & 67.92  \\
    LongLLMLingua & 6.91  & 34.74  & 12.56  & 31.86  & 63.94  \\
    LLMLingua2 & 16.50  & 48.21  & 24.00  & 44.53  & 72.14  \\
    \textbf{PartPrompt} & \textbf{19.80 } & \textbf{48.34 } & \textbf{26.60 } & \textbf{45.21 } & \textbf{72.59 } \\
    \bottomrule
    \end{tabular}
  \label{LLMBase}
\end{table}

\subsection{Ablation Study of PartPrompt}
\label{AblationStudy}

\begin{figure}[tp]
    \centering
    \includegraphics[width=0.46\textwidth]{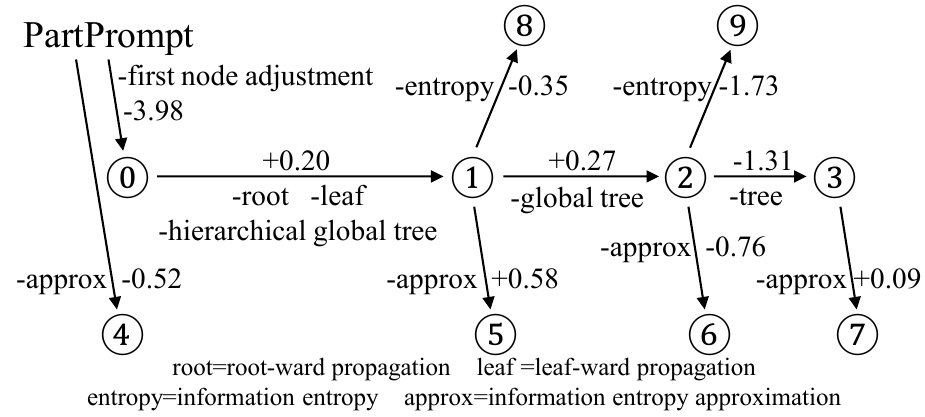}
    \caption{
    The organization of variants of PartPrompt in the ablation study. Each arrow comes with the removed component(s) and the corresponding performance drop (if "$-$" or gain if "$+$"). Each circled number \textcircled{0}-\textcircled{9} at the end of an arrow indicates the variant of PartPrompt after removing the component of that arrow based on the variant at the beginning of that arrow. The detailed performance of these variants can be found in Table \ref{alabation_perf}.
    }
    \label{ablationtree}
\end{figure}

To elucidate the effect of each PartPrompt component, a comprehensive ablation study is carried out. A total of ten variants of PartPrompt are designed and evaluated, and their organization is illustrated by a tree in Figure \ref{ablationtree}.

The complete PartPrompt is used as a point of departure for the sequential removal of components, so as to observe the effect of each component.
Initially, the first node adjustment technique is ablated, indicating that hyper-parameter $a_2$ was constrained to 1, noted as variant \textcircled{0}. Then the whole value adjustment module is omitted, resulting in the hierarchical global tree degenerating into a global tree with a single virtual node. This modification is referred variant \textcircled{1}. 
Subsequently, the global parse tree is removed, keeping only the local parse trees, which is denoted as variant \textcircled{2}. 
In this instance, each local parse tree is pruned independently and then merged.  
Finally, the local parse trees are removed, with only information entropy used for prompt compression, which is denoted as variant \textcircled{3}. Note that, the complete PartPrompt and its variants \textcircled{1}, \textcircled{2}, and \textcircled{3} all employ local approximated information entropy. When the information entropy approximation module is further removed, the complete PartPrompt and the corresponding variants \textcircled{1}, \textcircled{2}, \textcircled{3} become \textcircled{4}, \textcircled{5}, \textcircled{6}, \textcircled{7} respectively. 
These four variants can also be considered as an ablation chain of the global tree, and the performance difference among them can also reflect the contributions of each  component of global tree.

On the other hand, the variants \textcircled{8} and \textcircled{9} are developed to investigate the impact of entirely removing the information entropy module and solely utilizing the parse tree for compression.
Regarding a compression ratio of 0.5 on BBCnews, the performance of the complete PartPrompt, all ten variants, as well as four baseline methods is shown in Table \ref{alabation_perf}.

According to Table \ref{alabation_perf}, the complete PartPrompt achieves the best performance among all variants, demonstrating the effectiveness of each component. Notably, variants \textcircled{8} and \textcircled{9}, which rely solely on the parse tree, also achieve commendable results. The variant \textcircled{8} even outperforms three baseline selective methods: Selective-Context, LLMLingua and LongLLMLingua. These observations imply that the parse tree, analyzed by linguistic rules, holds valuable information for prompt compression.
Furthermore, the performance of variants \textcircled{4}, \textcircled{5}, \textcircled{6}, and \textcircled{7} decreases as the corresponding component removes, which also reflects their positive contributions.

Figure \ref{ablationtree} annotates the rise and drop of BS-F1 scores in the ablation study.  
First, it is evident that the first node adjustment (comparing to the variant \textcircled{0}) yields a significant performance improvement (3.98\%), 
which verifies the significance of considering human writing logic in prompt compression.
Second, variants \textcircled{8} and \textcircled{9} consider the parse tree, while variants \textcircled{1} and \textcircled{2} consider both the parse tree and information entropy and thereby obtain better performance (0.35\% and 1.73\%) respectively. This observation indicates that the parse tree and the information entropy are complementary to each other in prompt compression.

Section \ref{TheoreticalAnalysis} established that the information entropy approximation introduces an inter-sentence error, which can cause performance degradation if not managed. This error manifests clearly in our ablation results. For instance, variant \textcircled{3} (which uses approximation) underperforms compared to variant \textcircled{6} (which does not).
Furthermore, variant \textcircled{2}'s design inherently mitigates this issue. By performing compression within individual sentences, it avoids comparing token values across sentences, thereby eliminating the source of the inter-sentence error entirely. This explains its strong performance.
In contrast, variant \textcircled{1}'s reliance on a global parse tree necessitates cross-sentence token comparisons. Consequently, it is susceptible to the inter-sentence error introduced by the approximation, leading to the observed performance gap between variants \textcircled{1} and \textcircled{2}, and also between \textcircled{1} and \textcircled{5} (the non-approximated counterpart of \textcircled{1}).
The performance of variant \textcircled{0} further indicates that root propagation and leaf propagation alone are insufficient to correct errors and must be used together with first node adjustment.
Crucially, the complete PartPrompt architecture addresses this challenge. While it employs the global tree like variant \textcircled{1} (and thus faces potential inter-sentence error), it incorporates the node value adjustment mechanism specifically to correct the approximation bias, which achieving the best overall performance.

\begin{table}[tp]
  \centering
  \caption{
  The performance of variants of PartPrompt in ablation study. The circled numbers \textcircled{1}-\textcircled{9} denote the variants following their organization as shown in Figure \ref{ablationtree}.
  }
    \begin{tabular}{>{\centering\arraybackslash}m{2cm}
    >{\centering\arraybackslash}m{0.7cm}
    >{\centering\arraybackslash}m{0.85cm}
    >{\centering\arraybackslash}m{0.85cm}
    >{\centering\arraybackslash}m{0.85cm}
    >{\centering\arraybackslash}m{0.75cm}
}
    \toprule
    methods & BLEU & Rouge1 & Rouge2 & RougeL & BS-F1 \\
    \midrule
    \textbf{PartPrompt} & \textbf{28.28 } & \textbf{56.48 } & \textbf{38.30 } & \textbf{54.29 } & \textbf{74.57 } \\
    \midrule
    \textcircled{0}  & 18.78  & 49.65  & 27.43  & 46.82  & 70.59  \\
    \textcircled{1}  & 17.08  & 49.68  & 25.50  & 46.72  & 70.79  \\
    \textcircled{2}  & 18.10  & 50.21  & 26.29  & 47.30  & 71.06  \\
    \textcircled{3}  & 14.61  & 47.24  & 22.55  & 43.94  & 69.75  \\
    \midrule
    \textcircled{4}  & 28.20  & 55.95  & 37.86  & 53.77  & 74.04  \\
    \textcircled{5}  & 18.83  & 50.88  & 27.38  & 47.85  & 71.37  \\
    \textcircled{6}  & 16.63  & 48.95  & 24.86  & 45.91  & 70.29  \\
    \textcircled{7}  & 16.12  & 47.74  & 23.93  & 44.66  & 69.83  \\
    \midrule
    \textcircled{8}  & 18.01  & 49.04  & 26.56  & 46.30  & 70.44  \\
    \textcircled{9}  & 15.44  & 46.94  & 23.82  & 44.18  & 69.33  \\
    \midrule
    Selective-Context & 14.26  & 46.74  & 22.63  & 43.41  & 69.53  \\
    LLMLingua & 15.71  & 45.98  & 23.21  & 43.17  & 69.52  \\
    LongLLMLingua &	14.65 &	44.31 &	21.27 &	41.33 &	68.22 \\
    LLMLingua2 & 17.65 &50.12 &	26.06 &	46.89 &	71.39 \\

    \bottomrule
    \end{tabular}
  \label{alabation_perf}
\end{table}

\subsection{The Extreme Long Prompt Scenario}
\label{InputLimitation}

PartPrompt as well as Selective-Context and LLMLingua, employs a small language model to calculate the information entropy of each token. When the original prompt is extreme long and exceeds the input limit of the small language model, both Selective-Context and LLMLingua become impractical as they computes the information entropy over the whole prompt. In contrast, PartPrompt calculates the information entropy within each sentence, thereby enabling it to manage the extreme long prompt with many sentences.

To demonstrate the capacity of PartPrompt in processing the extreme long prompt, we select the papers, exceeding 6000 tokens, from the arXiv dataset without truncation. Since the length far exceeds the input limit of the small language model for calculating entropy, Selective-Context and LLMLingua (two state-of-the-art selective prompt compression methods) are no longer applicable. To this end, we employ Qwen2-72B to directly generate compressed prompt for comparison. 
It is important to note that using an LLM for compression requires a significant amount of computational resources, which may not be feasible in practical scenarios. The experimental results are shown in Table \ref{inputlimitresulttable}, and the abstract written by humans is employed as the ground truth.

\begin{table}[htbp]
  \centering
  \caption{The performance of PartPrompt for the extreme long prompt scenario. Other selective compression methods cannot handle this scenario, so a generative LLM is used for comparison.}
    \begin{tabular}{>{\centering\arraybackslash}m{2cm}
    >{\centering\arraybackslash}m{0.7cm}
    >{\centering\arraybackslash}m{0.85cm}
    >{\centering\arraybackslash}m{0.85cm}
    >{\centering\arraybackslash}m{0.85cm}
    >{\centering\arraybackslash}m{0.75cm}
}
    \toprule
    methods & BLEU & Rouge1 & Rouge2 & RougeL & BS-F1 \\
    \midrule
    LLM Generation & 0.00  & 13.33  & 0.76  & 12.29  & \textbf{52.67 } \\
    PartPrompt & 0.00  & \textbf{14.37 } & \textbf{0.77 } & \textbf{13.41 } & 51.07  \\
    \bottomrule
    \end{tabular}
  \label{inputlimitresulttable}
\end{table}

It can be observed that PartPrompt outperforms LLM Generation (by Qwen2-72B) on more metrics, despite LLM Generation receives better performance on the BS-F1 metric. Besides, the BLEU metric for both methods is zero due to the extreme large compression request. Considering the first time to test the extreme long prompt scenario, achieving such results is particularly encouraging for the selective prompt compression methods in the literature. 

\label{Visualizationstudy}
\begin{figure*}[hbp]
    \centering
    \caption{
    The compression results of PartPrompt (without adjustment) and the comparative methods on a brief intuitive example. All retained words are highlighted in green. LLMLingua retains tokens with incomplete semantics. Selective-Context retains too many function words. PartPrompt without adjustment can effectively identify key information and retain it. PartPrompt's compressed prompt is more coherent.
    }
    \label{IntuitiveExamples}
    \includegraphics[width=0.95\textwidth]{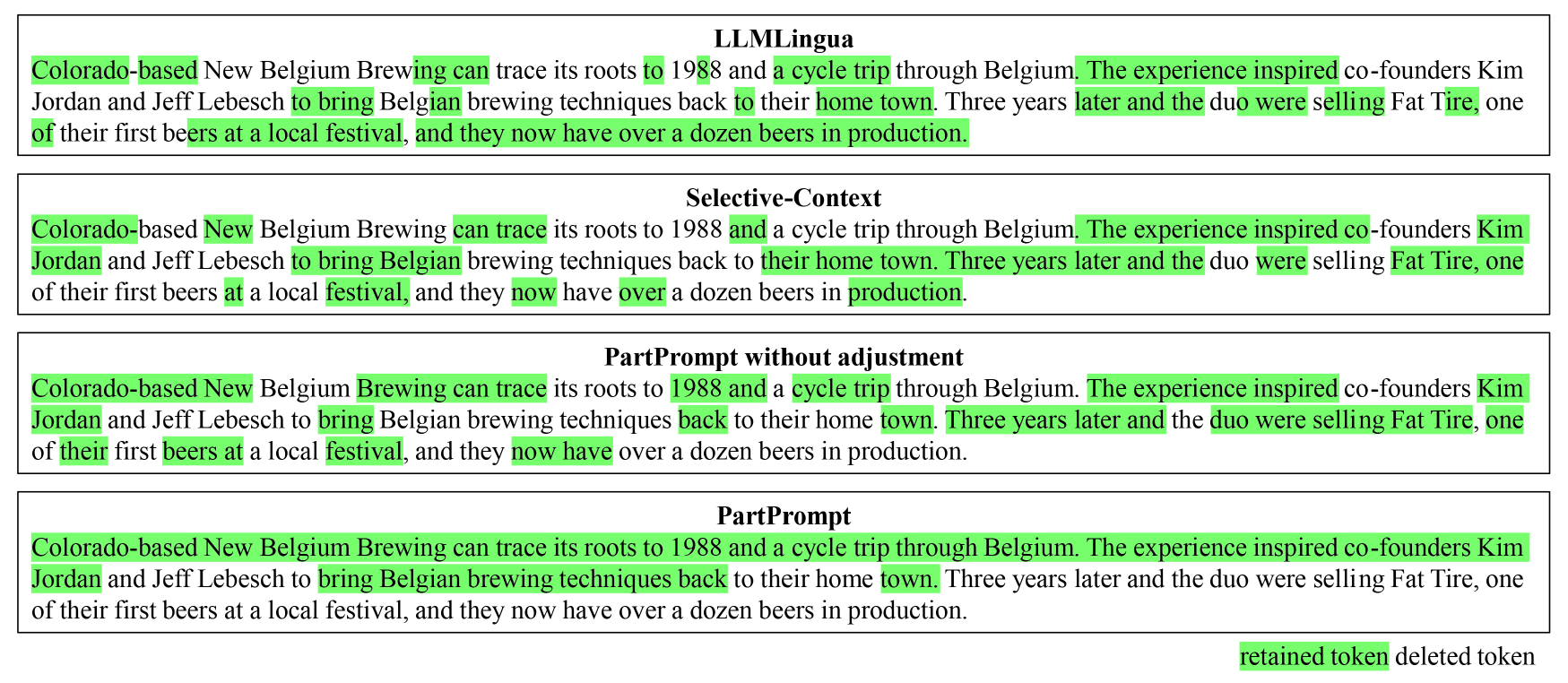}
\end{figure*}

\subsection{Non-discourse and Discourse Metrics}
\label{coherencestudy}
To analyze whether the compressed prompt can effectively preserve the original prompt while maintaining coherence, a direct comparison is made between the compressed prompt and the original prompt, with both non-discourse and discourse metrics employed for evaluation. 
Non-discourse metrics is employed to measure the similarity of texts at the lexical level, including BLEU, Rouge, and BERTScore. Concretely, Rouge (including Rouge1, Rouge2, and RougeL) measures the overlap between the hypothesis and reference texts. BERTScore (BS-F1 is used) offers a more comprehensive analysis of lexical similarity. 
BLEU consists of 1-gram, 2-gram, 3-gram, and 4-gram, where the n-gram calculates the proportion of n consecutive identical words. BLEU is employed to evaluate the continuity of the compressed text at the lexical level while measuring the lexical similarity. All the above non-discourse metrics are converted into percentages.

In contrast to non-discourse metrics, discourse metrics concern more on the global structure of texts, which can further evaluate the coherence of the compressed prompt \cite{grosz1995centering,mann1988rhetorical,zhao2022discoscore}. Specifically, RC and LC \cite{wong2012extending} measure the number and proportion of words that serve the connecting role. EntityGraph \cite{guinaudeau2013graph} evaluates text coherence through graphs. LexicalChain \cite{gong2015document} measures the overlap of lexical chains between the hypothesis and reference texts. DiscoScore (including DS-Focus and DS-SENT) measures the difference between the hypothesis and reference texts through their focus. We adopt the same settings as in \cite{zhao2022discoscore} for these non-discourse metrics. 

Both BERTScore and DiscoScore have a maximum text length of 512 tokens, the BBCnews dataset is thus truncated to 500 tokens in the experiments. The compression ratio is set to 0.5. The results are shown in Table \ref{coherence}, and note that the lower DS-Focus indicates the better performance.
\begin{table}[htbp]
  \centering
  \caption{The performance of a direct comparison between the compressed prompt and the original prompt. The lower score is better with $\downarrow$ symbol, otherwise the higher score is better.}
\begin{tabular}{>{\centering\arraybackslash}m{1.7cm}
    >{\centering\arraybackslash}m{2cm}
    >{\centering\arraybackslash}m{1.6cm}
    >{\centering\arraybackslash}m{1.6cm}
}
    \toprule
    metrics & Selective-Context & LLMLingua & \textbf{PartPrompt} \\
    \midrule
    Rouge1 & 66.20  & 63.58  & \textbf{74.80 } \\
    Rouge2 & 33.50  & 45.10  & \textbf{59.96 } \\
    RougeL & 65.37  & 62.80  & \textbf{74.68 } \\
    BS-F1 & 61.76  & 67.90  & \textbf{78.29 } \\
    \midrule
    BLEU 1-gram & 31.26  & 37.04  & \textbf{37.15 } \\
    BLEU 2-gram & 16.40  & 28.99  & \textbf{31.73 } \\
    BLEU 3-gram & 8.85  & 25.92  & \textbf{28.28 } \\
    BLEU 4-gram & 4.99  & 24.63  & \textbf{25.77 } \\
    \midrule
    RC & 0.378  & 0.388  & \textbf{0.607 } \\
    LC & 0.769  & 0.891  & \textbf{2.426 } \\
    EntityGraph & 0.402  & 0.310  & \textbf{0.480 } \\
    LexicalChain & 0.159  & 0.128  & \textbf{0.190 } \\
    DS-Focus $\downarrow$ & 1.262  & 1.050  & \textbf{1.049 } \\
    DS-SENT & 0.762  & \textbf{0.882 } & 0.869  \\
    \bottomrule
    \end{tabular}
   \label{coherence}
\end{table}

PartPrompt exhibits the superior performance in almost all metrics for both non-discourse and discourse metrics. In particular, PartPrompt considerably outperforms other methods for most discourse metrics, i.e., better at maintaining the global structure and coherence of the original prompt. This would not only enhance the effectiveness of the compressed prompt while feeding it to LLMs, but also facilitate humans to understand how and why prompt compression may work.

\subsection{Visualization of Contents Before and After Compression}
This section intends to provide intuitive examples of the prompts before and after compression. We extract a segment from BBCnews and compare four selective prompt compression methods: Selective-Context, LLMLingua, PartPrompt, and PartPrompt without adjustment. The results are visualized in Figure \ref{IntuitiveExamples}, and the retained tokens are marked in green.

LLMLingua tends to preserve incomplete words, which may cause by the direct use of LLM tokenizer. For instance, it preserves \textit{'8'} in \textit{'1988'}, \textit{'o'} in \textit{'duo'}, and \textit{'elling'} in \textit{'selling'}. The incompleteness of these words would lead to a fragmented compressed prompt, which makes it less coherent and comprehensible. While Selective-Context employs a LLM tokenizer along with some semantic rules, which results in the better completeness of words comparing to LLMLingua.
Note that, both methods employ information entropy to determine the value of words, so that some less informative words might be retained. For example, Selective-Context retains \textit{'and'} in \textit{'and a cycle trip through Belgium'}, and \textit{'now over'} in \textit{'they now have over a dozen beers'}. Besides, many function words alone, such as \textit{'at'} and \textit{'over'}, cannot offer meaningful information for LLM inference and human comprehension.

The variant of PartPrompt without node value adjustment is better able to preserve the main contents of sentences and remove function words with the help of the parse tree. This allows the proposed method to preserve more useful information and specific contents given the same compression ratio. After adding the node value adjustment module, PartPrompt can be regulated to preserve the first part and keep the completeness of the compressed prompt by adjusting its hyper-parameters. The resulting example by PartPrompt clearly obeys the desired goal, and looks more coherent and easier to understand than other resulting examples by other methods.

\section{Conclusion}
In this work, we introduced a novel prompt compression method called PartPrompt, which leverages parse trees to guide the prompt compressing process. To this end, the prompt compression problem was transformed into a tree pruning problem. And during the prompt compression process, the linguistic rules and human writing logic were achieved via constructing the global parse tree and adjusting the node value in the tree. The comparative experiments demonstrated the state-of-the-art performance of PartPrompt across various datasets (including different prompt task types and text types), compression ratios, metrics, and target LLMs for inference. A comprehensive ablation study verified the usefulness of each component in PartPrompt. And additional experiments further confirmed the advantages of PartPrompt in terms of more scenarios and metrics such as the extreme long prompt case and the coherence of compressed prompts.

For future work, although PartPrompt has been shown capable of handling extremely long prompts with encouraging results, further efforts are still needed to enhance the performance of current prompt compression methods under the extreme long prompt scenario. Besides, the patterns of LLM prompt analyzed by linguistic parse trees and human writing logic have been shown effective in improving the performance for the prompt compression problem, it remains unknown whether other patterns of LLM prompt are also helpful, which is another promising future direction.

\section*{Acknowledgements}
We would like to sincerely thank the anonymous reviewers and editors for their valuable advice. This work was funded by the National Key Research and Development Program of China (grant no. 2022YFF1202400), and was also supported by the Henan Province Medical Science and Technology Research Plan Major Project jointly built by the province and the ministry (grant no. SBGJ202401001).

\bibliographystyle{IEEEtran}
\bibliography{IEEEtran.bib}

\begin{comment}
\end{comment}

\vfill

\section*{Supplementary Materials \\ to the Paper \\ "Parse Trees Guided LLM Prompt Compression"}

\appendices 

\section{Detailed Complexity Analysis} 
\label{Detailed Complexity Analysis}

We provide detailed explanations for some conclusions in Section F here.
To keep the exposition clear, we restate the notation: 

$m$: number of sentences in the input prompt $R$.

$n$: number of tokens in each sentence (all sentences are assumed to have the same length).

$r_{j,i}$: the i-th token of the j-th sentence.

$C(\cdot)$: length function, so $C(R)=mn$.

$K$: a constant related to the model size used for entropy computation.
Since PartPrompt and the three baseline methods all rely on models with 7B parameters, the constant K can be considered identical for all four methods.

Selective-context uses a small model to directly compute the information entropy of each token in the input prompt. Since the computational cost of a transformer model is proportional to the square of the input text length, we have $\mathcal{O}(\text{Selective-Context})=\mathcal{O}(Km^2n^2)$,

LLMLingua splits the input prompt $R$ into a list of sentences $[R_1,R_2,\cdots,R_m]$. For the tokens in the j-th sentence, the information entropy of each token is computed using the following method:
\begin{equation}
    E(r_{j,i})=-\log~p_{\text{LM}}(~r_{j,i}~|~r_{j,<i},r_{<j,\text{retained}}).
\end{equation}
That is, for sentences $R_{<j}$, their compressed versions are used. 

Again, considering the quadratic complexity of the transformer, together with the fact that part of the KV cache can be reused, and additionally assuming a uniform compression ratio $\tau$ for every sentence, we obtain
$\mathcal{O}(\text{LLMLingua})=\mathcal{O}(Kmn^2(1+(m-1)\tau+(m-2)\tau^2))$.

Next, we provide a detailed analysis of the computational complexity of Algorithm 2: Recursive Tree Compression (corresponding to Line 8 in Algorithm 3). 

Lines 2-7 of the algorithm are used to merge two solution lists. Line 3 iterates over $l$, and line 4 in fact iterates over $z$. It follows that for every combination $(i,j)$, where $0\leq i\leq C(Q)-1$ and $0\leq j\leq C(P)-1$, the pair $(Q_i, P_j)$ is traversed exactly once. Thus, the complexity of the function \texttt{MergeSolution} is $\mathcal{O}(C(Q)\cdot C(P))$.

Next, we consider the complexity of lines 11 and 13 of the algorithm for an input node $v_i$. Line 11 iterates over all children of $v_i$. Suppose $v_i$ has $b$ child nodes, denoted as $v^{(i)}_1, v^{(i)}_2, ..., v^{(i)}_k, ..., v^{(i)}_b$. Let the length of the subtree corresponding to child node $v^{(i)}_k$ be $C(V^{(i)}_k)$. The length of the subtree rooted at $v_i$ is $C(V_i) = \sum_{j=1}^b C(V^{(i)}_j) + C(v_i)$.

For a specific child node $v^{(i)}_k$, after executing Line 12, we have $C(P) = C(V^{(i)}_k)$ and $C(Q) = \sum_{j=1}^{k-1} C(V^{(i)}_j)$. Consequently, the computational cost of Line 13 is $C(Q)\cdot C(P)=C(V^{(i)}_k)\cdot\sum_{j=1}^{k-1} C(V^{(i)}_j)$. Summing this cost over the iteration in Line 11 (over all children $v^{(i)}_k$), the total computational cost incurred by Line 13 during the traversal in Line 11 is:
\begin{align}
&\sum_{k=1}^b\left(C(V^{(i)}_k)\cdot\sum_{j=1}^{k-1} C(V^{(i)}_j)\right)\\
=&\sum_{k=1}^b\sum_{j=1}^{k-1}\left(C(V^{(i)}_k) C(V^{(i)}_j)\right)\\
=&\frac{1}{2}\left(\left(\sum_{k=1}^b C(V^{(i)}_k)\right)^2-\sum_{k=1}^b\left(C(V^{(i)}_k\right)^2\right)\\
\leq&\frac{1}{2}\left(\left(\sum_{k=1}^b C(V^{(i)}_k)\right)^2-\frac{1}{b}\left(\sum_{k=1}^b C(V^{(i)}_k)\right)^2\right)\\
=&\left(\frac{b-1}{2b}\right)\left(C(V_i)-C(v_i)\right)^2\\
\leq&\left(\frac{b-1}{2b}\right)\left(C(V_i)\right)^2
\end{align}

Therefore, when $C(V_i)$ is fixed, the computational cost incurred by Line 13 does not exceed $\left(\frac{b-1}{2b}\right)\left(C(V_i)\right)^2$. Equality holds when $C(V^{(i)}_j)=C(V^{(i)}_k),\forall j,k$ and $ C(v_i)=0$. Furthermore, the computational cost of Lines 9-10 and 14-20 is $\mathcal{O}(1)$ and $\mathcal{O}(C(V_i))$, which is negligible. Since the algorithm is recursive, the computation required by Line 12 is essentially performed within the "Line 13" corresponding to another node. Consequently, the overall computational cost of the algorithm is primarily attributed to the "Line 13" computation for each node in the given tree. This cost is maximized when the subtree lengths of all child nodes are equal. Therefore, the total computational cost of the algorithm is maximized when the given tree is a balanced tree. 

With the above analysis, we can calculate the complexity of the entire algorithm. Consider a balanced tree $T=(R,V,v_\text{root},f)$  with $l$ levels. That is, for any leaf node $v_i,f^{l-1}(v_i)=v_{root}$. All leaf nodes have a length of 1, while all other nodes have a length of 0 and $b$ children, $b \geq 2$. This configuration represents the worst-case scenario for computational cost. Denoting all leaf nodes as level 1, their parents as level 2, grandparents as level 3, etc. Then, level $k$ contains $b^{l-k}$ nodes, each node at level k corresponding to a subtree with length $b^{k-1}$. Therefore, the computational cost associated with "Line 13" for all nodes at level $k$ does not exceed$\left(\frac{b-1}{2b}\right)\left(b^{k-1}\right)^2b^{l-k}$. 
Hence, the total computational cost of the algorithm, i.e., the sum of the "Line 13" costs across all nodes in all levels of the tree, $\mathcal{O}(\text{Recursive Tree Compression})$, is:
\begin{align}
=&\sum_{k=1}^{l}\left(\frac{b-1}{2b}\right)\left(b^{k-1}\right)^2b^{l-k}\\
=&\frac{1}{2}\left(b^{2l-2}-b^{l-2}\right)\\
<&\frac{b^{2l-2}}{2}
\end{align}

The total length of all nodes in the tree $C(V)=b^{l-1}$. It follows directly that:
\begin{align}
&\mathcal{O}(Recursive Tree Compression)\\
&\leq\mathcal{O}\left(C(V)^2\right)=\mathcal{O}(m^2n^2)
\end{align}

\section{Experiments on hotpotQA and GSM8K dataset}
To test the performance of the proposed method on more diverse prompts, the hotpotQA\cite{yang-etal-2018-hotpotqa} and GSM8K\cite{cobbe2021trainingverifierssolvemath} datasets are also employed for additional testing. The answers in the dataset itself are used as ground truth. For hotpotQA dataset, we follow \cite{li2023leveraging} to compute precision, recall, and F1 in addtional, since the answers is quite brief. For GSM8K dataset, as the questions are math-related, we adopt exact matching to calculate the EM score obeying the same setting as in \cite{jiang2023llmlingua,fu2023chainofthoughthubcontinuouseffort}. 
The related hyper-parameters are set as follows: for hotpotQA, $a_1=3$, $a_2=20$; for GSM8K, $a_1=5$, $a_2=25$. The experimental results are shown in Table \ref{hotpotQAexperiment}.
It is evident that PartPrompt achieves remarkable improvements in almost compression ratios, and metrics in both datasets, which demonstrates the advantages of PartPrompt.

\begin{table*}[htbp]

\centering
\caption{
The performance of PartPrompt in comparison with baseline methods on hotpotQA and GSM8K datasets. 
}
\renewcommand{\arraystretch}{1.15}
\setlength{\aboverulesep}{0pt}
\setlength{\belowrulesep}{0pt}
\scalebox{0.96}{
\begin{tabular}{>{\centering}m{2cm}
>{\centering}m{0.7cm}
>{\centering}m{0.85cm}
>{\centering}m{0.85cm}
>{\centering}m{0.85cm}
>{\centering}m{0.75cm}
>{\centering}m{0.85cm}
>{\centering}m{0.7cm}
>{\centering}m{0.7cm}
>{\centering}m{0.8cm}
>{\centering}m{0.75cm}|
>{\centering}m{0.7cm}
>{\centering}m{0.8cm}
>{\centering\arraybackslash}m{0.75cm}
}
\toprule
\multirow{2}[2]{*}{methods} & \multicolumn{10}{c|}{hotpotQA}&\multicolumn{3}{c}{GSM8K} \\
   & BLEU & Rouge1 & Rouge2 & RougeL & BS F1 & precsion & recall & F1 & Tokens  & $1/\tau$ & EM & Tokens  & $1/\tau$ \\
\hline
LLM Generation&1.54&31.02&16.58&30.89&51.29&16.51&57.89&22.12&610.5&2.55&38.51&477.5&1.89\\
\hline
\multicolumn{14}{c}{20\% tokens constraint} \\
\hline
Selective-Context & 0.43  & 20.74  & 10.47  & 20.62  & 45.63  & 9.92  & 37.51  & 13.79  & 309.7  & 5.02  & 0.00  & 261.4  & 3.45  \\
LLMLingua & 0.97  & 21.12  & 11.44  & 21.06  & 47.28  & 10.42  & 37.00  & 14.26  & 309.5  & 5.02  & 3.03  & 155.2  & 5.81  \\
\textbf{PartPrompt} & \textbf{1.15 } & \textbf{24.62 } & \textbf{13.04 } & \textbf{24.57 } & \textbf{48.68 } & \textbf{12.61 } & \textbf{42.31 } & \textbf{17.13 } & 303.9  & 5.11  & \textbf{23.88 } & 179.4  & 5.01  \\
\hline
\multicolumn{14}{c}{30\% tokens constraint} \\
\hline
Selective-Context & 0.73  & 21.15  & 11.26  & 21.05  & 46.48  & 10.70  & 39.77  & 14.63  & 486.9  & 3.19  & 0.23  & 345.4  & 2.61  \\
LLMLingua & 0.65  & 21.24  & 10.57  & 21.15  & 46.21  & 10.56  & 37.63  & 14.31  & 453.2  & 3.43  & \textbf{67.85 } & 298.2  & 3.02  \\
\textbf{PartPrompt} & \textbf{1.96 } & \textbf{28.18 } & \textbf{15.95 } & \textbf{28.16 } & \textbf{50.17 } & \textbf{14.88 } & \textbf{48.04 } & \textbf{19.91 } & 456.8  & 3.40  & 53.30  & 270.4  & 3.33  \\
\hline
\multicolumn{14}{c}{50\% tokens constraint} \\
\hline
Selective-Context & 1.25  & 25.63  & 12.75  & 25.60  & 48.76  & 13.25  & 47.39  & 17.78  & 834.3  & 1.86  & 0.08  & 524.4  & 1.72  \\
LLMLingua & 0.87  & 21.99  & 10.97  & 21.86  & 46.88  & 10.70  & 40.83  & 14.74  & 764.4  & 2.03  & 12.74  & 425.5  & 2.12  \\
\textbf{PartPrompt} & \textbf{1.51 } & \textbf{27.89 } & \textbf{14.43 } & \textbf{27.85 } & \textbf{49.90 } & \textbf{14.05 } & \textbf{47.95 } & \textbf{19.10 } & 770.4  & 2.02  & \textbf{55.27 } & 446.4  & 2.01  \\
\hline
\end{tabular}
}
\scalebox{0.96}{
\begin{tabular}{>{\centering}m{2cm}
>{\centering}m{0.7cm}
>{\centering}m{0.8cm}
>{\centering}m{0.8cm}
>{\centering}m{0.8cm}
>{\centering}m{0.75cm}
>{\centering}m{0.75cm}
>{\centering}m{0.6cm}
>{\centering}m{0.7cm}
>{\centering}m{0.8cm}
>{\centering}m{0.8cm}
>{\centering}m{0.7cm}
>{\centering}m{0.8cm}
>{\centering\arraybackslash}m{0.8cm}
}
\hline
&&&&&&&&&&&&&
\end{tabular}
}

\label{hotpotQAexperiment}
\end{table*}

\section{Experiments on long prompt dataset}
Many advanced LLMs work with longer prompts, making it necessary to evaluate the proposed method’s performance on longer prompts.
We select the niah subset of RULER \cite{hsieh2024ruler} and LongBench \cite{bai2023longbench} datasets as benchmarks, sampling 100 test questions from each. The answers in the dataset itself are used as the ground truth. Due to the inherent length limitations of Selective-Context and LLMLingua, we therefore only compare our method against LLMLingua2 and LLM generative compression. 
For RULER-niah, we additionally report the performance of uncompressed prompts; this is infeasible for LongBench due to its excessive length. The hyper-parameters are set to $a_1=1, a_2=1$ for RULER-niah and $a_1=1, a_2=10$ for LongBench.

The experimental results are shown in the Table \ref{longbench}.
It can be observed that on the RULER-niah dataset, PartPrompt significantly outperforms both baseline methods. Notably, both selective compression methods surpass the performance of the original uncompressed prompt, whereas LLM generation performs poorly. In fact, "niah" stands for "Needle-in-a-Haystack". This dataset is designed to test a model’s ability to retrieve critical information (the "needle") from long sequences cluttered with distracting text (the "haystack").
Therefore, a good compression method that can remove the distracting text while preserving the critical information simplifies the task, thereby enhancing the downstream model's score. The experiments demonstrate that PartPrompt excels at this. In contrast, the uncompressed prompt’s excessive noise inherently challenges the LLM’s retrieval capability, explaining the poor performance of LLM generation compression.
On the LongBench dataset, the proposed method also outperforms the compared methods, demonstrating its superiority.

\begin{table}[htbp]
\centering
  \caption{Performance of PartPrompt and compared methods on two longer datasets. Each dataset lists the scores and the length of the compressed prompt.}
    \begin{tabular}{ccccc}
    \toprule
    \multirow{2}[2]{*}{method} & \multicolumn{2}{c}{RULER} & \multicolumn{2}{c}{LongBench} \\
\cmidrule{2-5}       & score & length & score & length \\
    \midrule
    LLM Generation & 0.0  & 2760.8  & 26.0  & 1918.2  \\
    LLMLingua2 & 24.0  & 2776.3  & 20.7  & 1972.3  \\
    PartPrompt & \textbf{69.0} & \textbf{2150.0} & \textbf{26.4} & \textbf{1751.0} \\
    Original prompt & 21.0  & 19191.9  &  NA  & NA \\
    \bottomrule
    \end{tabular}
  \label{longbench}
\end{table}

\section{Experiments on entropy model}
\begin{figure*}[htbp]
    \centering
    \includegraphics[width=0.98\textwidth]{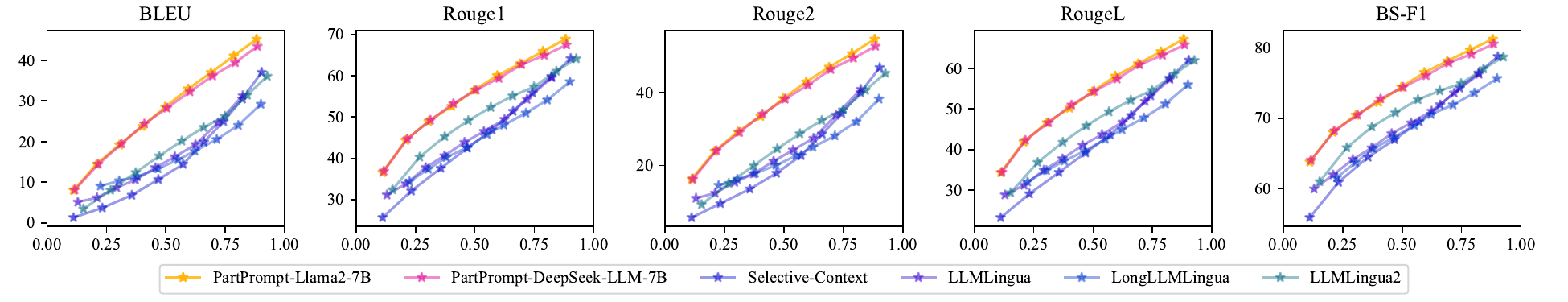}
    \caption{
    The performance of PartPrompt with another information entropy calculate model, DeepSeek-LLM-7B.
    }
    \label{deepseek}
\end{figure*}

The PartPrompt method employs a small model to calculate the information entropy of each token. In previous experiments, this is achieved using the Llama2-7B model. Considering that the model is an important part of the method, we show here a replacement experiment in which DeepSeek-LLM-7B\cite{bi2024deepseek} is employed to calculate the information entropy. We select the BBCnews dataset as the example, and the rest of the settings are the same as in Section VI-B.
The results of the experiment can be found in Figure \ref{deepseek}.

According to Figure \ref{deepseek}, the proposed PartPrompt achieves nearly identical performance while using different information entropy models, and both versions of PartPrompt consistently outperform the baselines. These observations demonstrate the robustness of our method in terms of using different information entropy computation models. 

% \bibliographystyle{IEEEtran}
% \bibliography{IEEEtran.bib}

\vfill

\end{document}